%% file: ms.tex
\documentclass[phd, titlesmallcaps, copyrightpage, foronline, openany]{mqthesis}

 \pdfoutput=1

\usepackage{rotating}  
\usepackage{theorem}   
\usepackage{bm}  
\usepackage{booktabs}  
\usepackage{pdfsync}   
\usepackage{bbm}
\usepackage{algpseudocode}
\usepackage{algorithm}

\usepackage[left=2cm,top=3.5cm,right=2cm,bottom=3cm,bindingoffset=0cm]{geometry}
\usepackage{bbm}
\usepackage{amsmath, amsfonts}

\usepackage{multirow}

\usepackage[protrusion=true,expansion=true]{microtype}

\usepackage[T1]{fontenc}
\usepackage[sc,osf]{mathpazo}  
\ifpdf
    \pdfcatalog { /PageMode (/UseOutlines) /OpenAction (fitbh)  }
\fi

\renewcommand{\P}[1]{P\left({#1}\right)}

\newcommand{\Psubhat}[2]{\hat{P}_{#1}\left({#2}\right)}
\newcommand{\Ptilde}[1]{\tilde{P}\left({#1}\right)}
\newcommand{\ind}[1]{\mathbbm{1}\left\{{#1}\right\}}
\newcommand{\fact}[3]{\texttt{#1(#2,#3)}}

\newcommand{\sw}{\textsc{SmallWorld} (Appendix~\ref{chap:sml})}

\renewcommand{\O}[0]{\mathcal{O}}
\newcommand{\A}[0]{\mathcal{A}}
\newcommand{\T}[0]{\mathcal{T}}
\newcommand{\R}[0]{\mathcal{R}}

\newcommand\numberthis{\addtocounter{equation}{1}\tag{\theequation}}

\newcommand*{\defeq}{\mathrel{\vcenter{\baselineskip0.5ex \lineskiplimit0pt
                    \hbox{\scriptsize.}\hbox{\scriptsize.}}}%
                     =}

\newtheorem{defn}{Definition}

\begin{document}

\frontmatter

\title{Probabilistic Models of Relational Implication}

\ifthenelse{\boolean{foronline}}{
  \author{\href{mailto:xavier.holt@students.mq.edu.au}{Xavier Ricketts Holt}}
  \department{Computing}
}{
  \author{Xavier Ricketts Holt}
  \department{Computing}
}

\degrees{BSc (First Class Honours) University of Sydney}


\renewcommand{\degreetext}
{in partial fulfilment of the Degree of Masters of Research}

\titlepage

\input{acknowledge}

\input{abstract}

\tableofcontents
\listoffigures
\listoftables

\mainmatter

\input{ch_intro/chap_intro}

\input{ch_litreview/chap_litreview}

\input{ch_method/chap_method}

\input{ch_exp/chap_exp}

\input{ch_conclusion/chap_conclusion}

\appendix


\input{ch_appendix/chap_app}

\backmatter

\input{listofsymbols}


\bibliography{references}

\end{document}

%% file: acknowledge.tex
\chapter{Dedication}

I am immensely grateful for the guidance and knowledge provided to me by my supervisory team. Thank to Mark, for the instruction, the rigour and the whiteboard sessions. Thanks to Mark, Javad and Nate for the direction and advice via Edinburgh, and the Tuesday evening Skype sessions. To Hegler, thank you for the instruction and the conversation (and for the use of your Macbook).

To my family and friends, thank you for supporting me through this whole process -- in particular towards the end when things (as always) get a little stressful. To Atia, thank you for your love and your support, and for getting through this together. Joel -- as always, I appreciate your advice and your ongoing crusade against my em-dashes.

Finally, thank you to my former University, for not shutting off my access to the departmental CPUS.

%% file: abstract.tex
\chapter{Abstract}

Knowledge bases and relational data form a powerful ontological framework for representing world knowledge. Relational data in its most basic form is a static collection of known facts. However, by learning to infer and deduct additional information and structure, we can massively increase the expressibility, generality, and usefulness of the underlying data. One common form of inferential reasoning in knowledge bases is implication discovery. Here, by learning when one relation implies another, we can implicitly extend our knowledge representation. There are several existing models for relational implication, however we argue they are sufficiently motivated but not entirely principled. To this end, we define a formal probabilistic model of relational implication. By using estimators based on the empirical distribution of our dataset, we demonstrate that our model outperforms existing approaches. While previous work achieves a best score of 0.7812 AUC on an evaluatory dataset, our ProbE model improves this to 0.7915. Furthermore, we demonstrate that our model can be improved substantially through the use of link prediction models and dense latent representations of the underlying argument and relations. This variant, denoted ProbL, improves the state of the art on our evaluatoin dataset to 0.8143. In addition to developing a new framework and providing novel scores of relational implication, we provide two pragmatic resources to assist future research. First, we motivate and develop an improved crowd framework for constructing labelled datasets of relational implication. Using this, we reannotate and make public a dataset comprised of 17,848 instances of labelled relational implication. We demonstrate that precision (as evaluated by expert consensus with the crowd labels) on the resulting dataset improves from 53\% to 95\%. We also argue that current implementations of link prediction models are not sufficiently scalable or parametisable. We provide a highly optimised and parallelised framework for the development and hyperparameter tuning of link prediction models, along with an implementation of a number of existing approaches.

%% file: ch_intro/chap_intro.tex
\chapter{Introduction}

Relational data provides a structured and flexible representation of knowledge. An instance of relational data is known as a proposition, and it is comprised of a relation and a tuple of arguments. Notationally, we represent a proposition as \texttt{r$(\mathtt{a_1, a_2, a_2, \dots})$} where \texttt{r} indicates the relation and $\mathtt{a_i}$ corresponds to the slot at index $i$. For example, \fact{tutors-at}{Sam}{University}, or \fact{prime-minister-of}{Malcolm Turnbull}{Australia} are instances of relational data with two arguments.

In general, arguments are objects, entities or other `things'. Relations in turn indicate some specific connection between the constituent arguments. We say the relation is instantiated by the tuple of arguments. Relational data is an enormously broad category. All graph representations can be viewed through the lens of relational data, where nodes play the part of arguments and edges relations. It also backs the most common database storage model, a relational database. As such, improving knowledge representation in relational data has enormous business and institutional application. 

Our goal in this work is to develop models of relational inference. In particular, we seek to learn which relations imply other relations. In doing so we can extend the expressibility and representability of existing knowledge bases. We justify our project within the current literature; while existing models are structurally similar to measures of conditional probability, they are not formally defined as such. We argue that a formal probabilistic model is theoretically a strong fit for the types of inference we seek to capture. Furthermore, we demonstrate empirically that the models formulated using our probabilistic representation are empirically superior to all previous work. 

We define a number of research questions that we seek to address in the body of this work. Firstly, we wish to determine if a formal probabilistic model provides additional insight and leverage for the relational implication problem. We test this by developing ProbE, a model which is probabilistic, but uses the same form of data as existing work -- that is, it makes no use of link prediction models to infer missing data. Secondly, we wish to determine whether the use of link prediction models improve performance for relational implication. To this end, we define ProbL -- a probabilistic model that uses link prediction data. By comparing the results of this model to ProbE, we can evaluate whether the additional structure is beneficial. Finally, we wish to evaluate whether current labelled datasets are sufficient for the task. In order to evaluate models, the standard procedure is to use a labelled corpus. However, the corpus that is best in terms of coverage and form was trained using crowd workers. We wish to evaluate whether the resulting labels are of sufficient quality, and in the event of this not holding we seek to provide another evaluation framework.

\input{ch_intro/sec_format}

\input{ch_intro/sec_impl}

\input{ch_intro/sec_contributions}

\input{ch_intro/sec_outline}

%% file: ch_intro/sec_format.tex
\section{Problem Format and Input Data}

\begin{table}
\centering
\begin{tabular}{@{}lll@{}}
\toprule
Relation            & Subject        & Object             \\ \midrule
\texttt{studies-at} & \texttt{Jane}  & \texttt{Macquarie} \\
\texttt{studies-at} & \texttt{Sam}   & \texttt{Macquarie} \\
\texttt{taught-by}  & \texttt{Emily} & \texttt{Sam}       \\
\texttt{teaches}    & \texttt{Sam} & \texttt{Emily}       \\
\texttt{tutors-at}  & \texttt{Sam}   & \texttt{Macquarie} \\
\texttt{works-for}  & \texttt{Jacob}  & \texttt{Macquarie} \\ 
\texttt{works-for}  & \texttt{Sam}  & \texttt{Macquarie} \\ \bottomrule
\end{tabular}
\caption{\textsc{SmallWorld}: an illustrative set of relational data we refer to frequently in this thesis.}
\label{tab:smallworld_in}
\end{table}

Relational data is often characterised by the size of the instantiating argument tuple. In particular, relations instantiated by two arguments (binary relations) are extremely commonplace, and form the basis for most existing knowledge bases. In this thesis we deal predominately with binary relational data, and when the size of the argument-tuple is not otherwise specified this is what we refer to. 

We say a relation and argument-tuple are compatible if the resulting proposition is true. In general we assume that all argument-relational pairings are determinable, however some may be unknown. As such, `truth' here refers to an idealised underlying ground truth. 

For pedagogy, we introduce a toy relational dataset henceforth referred to as \textsc{SmallWorld} (table~\ref{tab:smallworld_in}). For ease of reference, we also include \textsc{Smallworld} in Appendix~\ref{chap:sml}.

Notationally, we use $\O$ to denote a corpus of relational data. For heavily structured relational data, this is generally a set -- either data is observed or it is not. However, it is also possible to scrape relational data directly from text. In this case, $\O$ corresponds to a multiset where propositions can occur multiple times.

Each element of $\mathcal{O}$ contains a relation $r$ and a tuple of arguments $(s,o)$. For binary data, we adopt the convention that the first argument-slot ($s$ in our example) is referred to as the subject. Similarly, the second argument slot ($o$) is referred to as the object. It is often notationally convenient to refer to the argument tuple collectively -- collapsing the argument pair into a single variable $t = (s,o)$. We use this notation frequently throughout the thesis, in particular when referring to the new models developed in Chapter~\ref{ch:meth}. This representation also has the benefit of seamlessly generalising to n-ary tuples -- $t$ simply denotes the general argument-tuple in all cases.

We denote $\R$ to be the set of relations and $\A$ the set of arguments. Furthermore, let $\T$ denote the set of argument-tuples observed in $\O$ -- that is, $\T = \{t \mid (r,t) \in \O\}$. Elements of $t = (s,o)$ are ordered, so $t = (s,o) $  and $t = (o,s)$ naturally refer to different argument-tuples.

For illustration, in \textsc{SmallWorld} we have:
\begin{itemize}
\item $\R = $ \{\texttt{studies-at, taught-by, teaches, tutors-at, works-for}\};
\item $\A = $\{\texttt{Jane, Macquarie, Sam, Emily, Jacob}\}; and
\item $\T = $\{\texttt{(Jane, Macquarie)}, \texttt{(Sam, Macquarie)}, \texttt{(Emily, Sam)}, \texttt{(Sam, Emily)}, \dots\}.
\end{itemize}

%% file: ch_intro/sec_impl.tex
\section{Relational Implication}
\label{sec:intro_impl}

Our understanding of the world is informed by powerful semantic constructs -- for example, synonyms, hypernyms and antonyms. We use these to generalise and extend our knowledge of the world -- if we know that an animal is a dog, we know it is a mammal. If we know it is a mammal, we know it is warm-blooded but that it is not an invertebrate. These rules are learned either formally or contextually. We develop a partial model of the world and use it to inform, characterise and define objects and concepts in reference to others.

This form of inferential logic is not present in say a database of known facts. If we've observed that $X$ is a dog, we may nevertheless report that $X$ is not a mammal if this was not present in our dataset. However, learning where this form of inference applies is enormously powerful. Discovering rules of relational inference is a way of encoding inferential logic into knowledge bases. Here, the task is to discover which relations imply others -- for example, the rule \fact{tutors-at}{}{} $\rightarrow$ \fact{works-at}{}{}. This is particularly important for systems that interact with people through natural language -- such as with automated personal assistants or navigational interfaces, due to the plethora of possible formulations of the same underlying concept.

Implication, also known as entailment or the material conditional in formal logic is a binary operator between two statements. Implication is represented with the $\rightarrow$. Formally, we define it as follows:

\begin{defn}
Implication: an implication $p\rightarrow q$ is logically equivalent to $\lnot (p \land \lnot q)$. In an implication, the left hand statement ($p$ in our case) is known as the antecedent, and the right hand statement ($q$) is the consequent.
\end{defn}

Within this body of work, we refer to implication within a relational framework -- that is, implications are of the form $p\rightarrow q$ where $p,q\in \R^2$.

We refer to \sw{}. Possible implication rules might be $\texttt{tutors-at} \rightarrow \texttt{works-for}$. This instance of relational implication is relatively unambiguous -- for all $t$ such that $\texttt{tutors-at}(t)$ (the antecedent) is true, we may deduce that $\texttt{works-for}(t)$ (the consequent) is also compatible. However, requiring this to hold for all argument-tuples turns out to be quite restrictive, especially when dealing with natural language. For example, perhaps \fact{tutors-at}{Sam}{Macquarie} is in fact indicating that \texttt{Sam} is a tutor, not affiliated with \texttt{Macquarie} but simply using the premises. In this case, \fact{works-for}{Sam}{Macquarie} would not hold. This issue is particularly relevant when we scrape relational data directly from collections of natural language -- in general, our extraction procedure is not capable of distinguishing subtle semantic differences such as these. This is only complicated when we consider that propositions taken in this manner can be untrue -- either because of an error in the scraping process, or because the source material was incorrect.

As such, in this work we adopt the reconceptualisation of implication used in inductive logic\footnote{This is a term used in the philosophical literature, and differs from induction in a mathematical proof sense. See: \url{https://plato.stanford.edu/entries/logic-inductive/}}. Here, instead of a simple boolean rule, we have degrees of belief in $p \rightarrow q$. We can interpret this probabilistically -- if $\P{p\rightarrow q}$ is high, then for all $t$ such that $p(t)$ holds, we believe with high probability that $q(t)$ is also true. Again going back to our example, we might say that \texttt{works-for} $\rightarrow$ \texttt{tutors-at} with high probability -- while still allowing for edge cases such as the one discussed above.

This is a powerful representation even for implications that do not occur near unanimously. For example, consider what we know of $X$ if we observe \fact{tutors-at}{X}{Macquarie}. In particular, what can we say about the relative likelihood of \texttt{studies-at}{X}{Macquarie}? Many people both tutor and study at a University. As such, we may expect \texttt{studies-at}{X}{Macquarie} to be more true than if we had not observed \fact{tutors-at}{X}{Macquarie}. However, this form of inference would not be expressible in a formal deductive framework -- in a deductive sense \texttt{tutors-at} $\not\rightarrow$ \texttt{studies-at}. However, from an inductive perspective we could still suggest that the likelihood of \texttt{tutors-at} $\rightarrow$ \texttt{studies-at} is greater than if the two relations were semantically unrelated.

With this in mind, we define the task of relational implication as follows:

\begin{defn}
Relational implication: for all $p,q \in \R\times\R$, assign a value to $p\rightarrow q$ that indicates the degree of belief in $q(t)$ given that we know $p(t)$.
\end{defn}

In particular, in this thesis we refer to relational implication where the argument tuple is the same for antecedent and consequent relation. That is, if there is strong evidence for $p\rightarrow q$ and $p(t)$ is true then we can infer on $q(t)$. However, we can not reason on $q(t')$ where $t'$ is some other argument-tuple. While certain implicational rules may fall into this category \fact{treats}{aspirin}{headaches} $\rightarrow$ \fact{treatable-using}{headaches}{medicine}, the task of relational implication is confounded by argument implication. As such, we do not address implications of this form. One exception we make is where the tuple is composed of the same arguments, but in reverse order. This is a natural. frequently occurring form of relational implication. For example \fact{reads}{X}{Y} $\rightarrow$ \fact{read-by}{Y}{X} clearly holds. We demonstrate later in this work that implications of this form can be handled within the same framework as standard implicatoinal rules.

\section{Link Prediction}
\label{sec:link_predict}

Similarly, determining whether a proposition is true can be viewed probabilistically. For example, relations, in general, can only be instantiated by compatible types of arguments. Say we had not observed the proposition \fact{presented-award-to}{X}{Y} but were nevertheless required to reason on its likelihood. If we were given that $X$ and $Y$ are people, we might assign a higher likelihood to the proposition than if we were of the belief that $X$ and $Y$ were other entity types. This general form of inference, whereby we reason about the relative likelihood of unseen facts is known as link prediction\footnote{The task is also referred to as knowledge-base completion.}. We define:

\begin{defn}
Link prediction: for all $r(t) \in \R\times (\A \times \A)$, provide a score proportional to the likelihood that $r$ and $t$ are compatible, i.e. $r(t)$ is true.
\end{defn}
\label{def:link_pred}

We note that link prediction models can assign likelihood to all possible argument-tuples, not just those argument-tuples which were observed in $\O$.

There are clear thematic connections between link prediction and relational implication. Both attempt to reason probabilistically about unknown statements. For link prediction, the task is simply to assign a likelihood to an unseen proposition. For relational implication, one seeks to learn a higher level semantic relation. There is hence some benefit in considering these two tasks together. In one direction, known implication rules can improve existing models of link prediction. If we observe $p(t)$, have inferred $p\rightarrow q$ with high probability, and are asked to reason about the likelihood of $q(t)$ we would be quite confident.

In the reverse direction, as we demonstrate, relational implication can be scored based on the proportion of true propositions that are consistent with the implicational hypothesis (i.e. for $p \rightarrow q$, tuples such that $p(t)$ is true and $q(t)$ is true). Link prediction models dramatically increases the range of tuples over which we can score as likely or unlikely for a given relation. As such, implication rules with few observed instantiating tuples can nevertheless be scored over a wide range of possible propositions. Furthermore, with relational data scraped from natural language, observed propositions may be noisy or incorrect. Link prediction models can also be used to assess the likelihood of propositions that were observed -- perhaps providing a more robust measure of likelihood than simple counts. In our thesis, we make the first such model of relational implication that uses link prediction. We hypothesise that by including additional information of likely but unseen propositions we can improve the inferences built upon these propositions. 

%% file: ch_intro/sec_contributions.tex
\section{Contributions}

We make three primary contributions to the task of relational implication. 

We derive and justify a formal model of relational implication grounded in probability theory. We first note that the `Cover' measure which is used frequently as a score of relational implication can be viewed as a measure of conditional probability. We extend this idea by motivating and providing a Bayesian net framework of relational implication aimed specifically at measuring the conditional likelihood of one relation given that another is true. The framework can be used with a number of different measures of probability, and our first model simply makes use of the empirical distribution function of our input facts. We demonstrate that this measure provides state of the art performance across several tasks.

Our approach is general and extensible, and our second major contribution is to use this framework to provide a score of relational implication that makes use of the recent progress that has been made in link prediction models. Specifically, instead of relying simply on the empirical distribution of our known facts, we infer other likelihoods through link prediction models, which are capable of assigning meaningful scores to out-of-scope facts. This allows us to score over not only tuples which we have observed to be true, but tuples which we have not observed but are nevertheless likely. We use a variety of link prediction methods to provide the base scores on top of which our conditional likelihood is built, and again obtain state of the art results on a number of different tasks. We also provide several variants of the model which allow for efficient computation, as the full link prediction model sums over all tuples and for some datasets or purposes this is too computationally expensive. Additionally, in order to train link prediction models at a scale suitable for the dataset over which we operate, we have developed and built a high-performance module for training link prediction tasks. We provide a training framework that includes a highly optimised batch generation function and several existing link prediction models, and make our code publicly available.

We also argue that there is currently no ideal dataset for the relational implication task. In particular, we demonstrate that expert annotators frequently disagree with the crowd decision when identifying the direction of implication -- that is, they may identify that there is an implicational relation between \fact{tutors-at}{X}{Y} and \fact{staff-member-at}{X}{Y}, but might incorrectly decide on the direction of implication. We argue this is an artefact of how instances were originally presented to the crowd workers, and to this end define a new crowd task specifically aimed at addressing the shortcomings of previous work. We use our reformulation of the task to extend and reannotate an existing relational implication dataset, extending the size of the dataset from 13,000 rows to 17,000 and increasing expert agreement on the label of implicational rows from 55\% to 95\%.

%% file: ch_intro/sec_outline.tex
\section{Outline}

We provide an overview of relevant current literature in Chapter~\ref{ch:lit}. We describe existing sources of relational data and the procedure for generating this data; current approaches used to score relational implication rules; as well as existing processes and datasets for the evaluation of implicational inference. We also describe link prediction models in general, and outline the specific approaches we use for comparison in the development of our statistical model.

In Chapter~\ref{ch:meth} we motivate and describe a formal probabilistic model of relational implication. We represent the implicational score as a sum-product between marginal and bivariate conditional distributions. Based on our estimator of these distributions, we define different models of relational implication. Our first model, denoted ProbE, simply uses the empirical distribution over the input triples as the estimator of densities. Our subsequent models ProbL and ProbEL both incorporate a link prediction score function to define an estimator for conditional distributions. ProbL scores over all input tuples, and as such is capable of accounting for likely but unseen inputs. ProbEL scores over observed relation argument-tuple pairs, and as such is more efficient when compared to ProbL. It is also calculated over the same tuples as ProbE, DIRT, Cover and BINC and as such allows for a clear measure of comparison.

We describe our empirical formulation and results in Chapter~\ref{ch:exp}. We first use expert annotators to evaluate an existing labelled dataset for relational implication. We demonstrate that there is a very high error rate when considering relational implication rules that only apply in one direction, which correspond to a large proportion of the rules labelled as implicational. To address this we define a new crowd task formulation and validation framework, and use this to re-annotate the positive instances of the original dataset. We use this re-annotated corpus to score a collection of existing relational implication models, as well as those defined in Chapter~\ref{ch:meth}. We find that all novel formulations defined in this thesis outperform existing methods.

Finally, in Chapter~\ref{ch:con} we re-frame the content of our work in light of the empirical results discussed in the previous chapter. We also discuss and explore possible avenues for future work.

%% file: ch_litreview/chap_litreview.tex
\chapter{Literature Review}
\label{ch:lit}

In this chapter we describe existing work that relates to this thesis. We describe the approaches used to generate relational data, as well as the specific datasets we draw upon throughout this work. We then describe existing measures of relational implication in terms of how they are calculated and the work that has been done on extending them. We cover how relational implication has been scored -- initially through downstream evaluation, and more recently with specific pre-annotated evaluation sets. Next, we discuss the problem of link prediction as it pertains to our task. We discuss a number of different link prediction models and describe the general framework used to train them. Finally we discuss common extensions to the relational implication task, such as an implicational or entailment graph which covers the transitive nature of implication. In the final section of this chapter, we demonstrate how the contributions we outline in the previous section were guided and motivated through specific existing gaps in the literature.

\input{ch_litreview/sec_data}

\input{ch_litreview/sec_scores}


\input{ch_litreview/sec_emb}

\input{ch_litreview/sec_eval}



\section{Framing and Concluding Remarks}

In this section we describe how our contributions, outlined in Chapter 1 fit into the existing literature.

Our world knowledge, and representations of this knowledge are often probabilistic, and information and belief is not binary but continuous. So too is implication -- \fact{born-in}{PERSON}{COUNTRY} does not imply \fact{citizen-of}{PERSON}{COUNTRY}, but it does provided positive evidence for the latter. This is evident in existing measures such as Dirt or Cover, which evaluate, for an implication rule the proportion of tuples consistent with the implication hypothesis. This evaluates to a number between zero and one, and can roughly be conceptualised as a degree of belief in the implicational hypothesis. However, it only resembles a probabilistic statement -- it is not formally defined as such. We argue that due to the probabilistic format of belief and implication that such a representation is valuable.

On this note, there has been a large amount of work done in the task of link prediction. Here, given a set of input facts, the goal is to assign likelihoods to triples not present in input. Doing so allows inference of missing facts, and intuitively seems like it could be useful in models of relational implication. However, to the best of our knowledge there has of yet been no application of link prediction models to the task. We provide one such model, and demonstrate its efficacy empirically. In addition, current evaluation of link prediction models is impacted by inter-model factors such as hyperparameter tuning and training framework. It is therefore useful to have an efficient framework for hyperparameter tuning on a range of models in a principled manner, and the code we make public as part of this thesis is a first pass at doing so.

Our final contribution, the new annotation procedure for crowd task formulation is in part due to a perceived gap in the literature, and in part due to experimental analysis outlined in Chapter 4. To the former, we argue that existing methods of crowd annotation do not represent the difficulty of the task. It is our experience that crowd work is fickle, and enormous care and subtly (along with trial and error) is often required. We believe that the current formulation does a great job at identifying unrelated relations, but that it does less well at capturing the directionality of implication rules. Our formulation is specifically aimed at this shortcoming.

%% file: ch_litreview/sec_data.tex
\section{Sources of Relational Data}

In this section, we identify existing datasets of relational inference that occur frequently in the literatures of relational implication and link prediction tasks. We divide them into fixed schema and flexible schema datasets. 

\subsection{Fixed Schemas}

\begin{table}
\centering
\begin{tabular}{@{}llll@{}}
\toprule
Dataset      & $|\A|$ & $|\R|$ & $|\O|$       \\ \midrule
FB15K  & 14,951 & 1,345 & 592,213 \\
WN18  & 40,943 & 18 & 151,442 \\
NYT + OpenIE   & 14,500 & 5,500 & 1,800,000 \\
Google Books  & 6,150,600  & 437,089 & 1,530,553,834       \\ \bottomrule
\end{tabular}
\caption{Table of Common Datasets in the Link Prediction and Relational Implication Literature.}
\label{tab:ds}
\end{table}

Fixed schema datasets are characterised by a predefined set of possible relations. They are generally managed or curated, though this curation may be crowdsourced. Fixed schema datasets have the benefit of being immediately interpretable, and in general may be of a higher quality than automatically derived sets of relational data. Within the context of this work, structured datasets are typically used in the evaluation of link prediction models.

\begin{description}
\item[Freebase: FB15k] Freebase is a large collaborative knowledge base. Freebase scraped from a wide variety of data sources, including Wikipedia and other domain specific sources of information. Additionally, users could submit facts and verify facts submitted by others. Freebase was structured into broad topics, best understood as an amalgamation of descriptive, possibly overlapping tags that conveyed the entity in question. FB15K \cite{ds:freebase} is a subset of Freebase that is one of two de facto datasets for evaluating link prediction models. 
\item[WordNet: WN18] The other de facto dataset for the task is WN18 \cite{ds:wordnet} -- WordNet. WordNet is a lexical database of the English language, comprised of words structured into semantically equivalent `synsets'. Although not traditionally viewed as such, WordNet has been used as a relational dataset. In this case, relations are the semantic constructs the link words, such as synonymy, hypernymy, antonymy and meronymy. Arguments are the words (technically, the synsets) themselves. For example, \fact{hyponym-of}{Eagle}{Bird} would correspond to a proposition where $r = $ \texttt{hyponym-of} and $t = $\texttt{(Eagle, Bird)}.
\end{description}

\subsection{Flexible Schemas}

\begin{figure}[h]
\centering
\includegraphics[width=15cm]{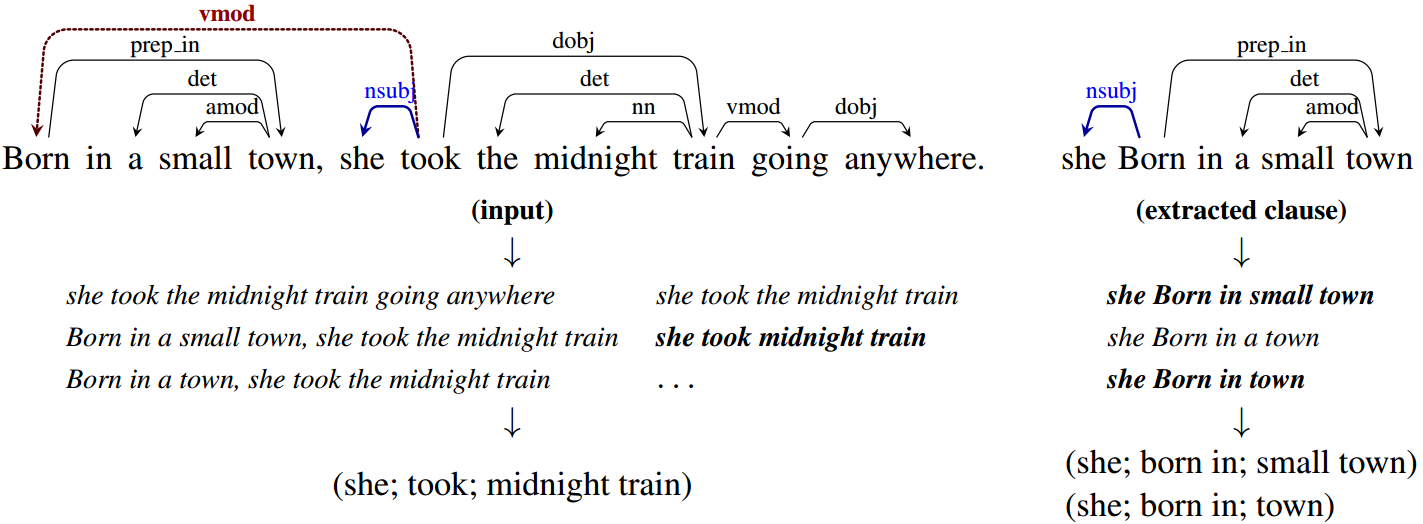}
\caption{OpenIE flexible schema approach. The underlying sentence is first parsed, and a number of entailed clauses matching a predefined syntactic pattern are identified. These clauses are then cleaned and shortened.}
\label{fig:open_ie}
\end{figure}

An alternative to using a predefined schema is to learn a relational representation from text. This method, characterised as flexible schema, takes as input a large corpus of unlabelled text. Each sentence is parsed to indicate, for each token, its part of speech as well as intra-token relations. Relational data follows a certain syntactic form, and these syntactic patterns are defined a-priori. Matching patterns in the parsed sentences are then identified, and often cleaned and shortened. For an example, see figure~\ref{fig:open_ie}.

\begin{description}
\item[OpenIE + NYT] One framework for relational extraction is OpenIE \cite{openie}. The system decomposes a sentence into entailed clauses before shortening the resulting parse tree and taking it as a relation in the schema. Riedel et al. \cite{riedel2013relation} used OpenIE on the New York Times dataset \cite{ds:nyt} to extract a corpus of 1,800,000 triples.
\item[Google Books] Another set of triples which has not received as much attention in the literature is the Google Books relational corpus extracted by Levy and Dagan \cite{Goldberg:2013wd}. Google books is a massive corpus of text created by scanning over 25 million titles and processing the resulting images with optical character recognition (OCR). Google provides a subset (of around 345 billion words) of this corpus tagged with syntactic parse information \cite{ngrams}. Levy and Dagan use these partial parse trees to generate a set of relational triples several orders of magnitude larger than anything else in the literature, at 1,530,553,834 rows and 64 billion unique propositions.
\end{description}

\subsection{Universal Schema}

Schemas can also be combined. Through entity-linking, the relations from one database can be matched against another. This allows for a larger set of relational data, and can improve performance on any specific dataset through transfer learning. Riedel et al. \cite{Riedel:2013ve} linked the NYT + OpenIE corpus to Freebase through simple entity-linking (string matching). In doing so, they found relations from the flexible-schema that could serve as excellent predictors hitherto unseen Freebase-relations. Taking the union of Freebase and a flexible schema also has the added benefit of improving the capability of a system to receive queries in natural language. Instead of having to know the strict Freebase relation-type, the coverage of NYT surface forms means that a user would be free to make a query whichever way they like (assuming that similar phrasing had been used somewhere in NYT).

%% file: ch_litreview/sec_scores.tex
\section{Scores of Relational Implication}
\label{sec:scores}

In this section we describe the methods currently being used in the literature to address the relational implication task. We separate them into three broad categories: set based measures, defined by simple overlaps; the entailment graph, where transitivity is used to formulate the problem as an integer linear program (ILP); and embedding methods, where dense relational embeddings are learned.

\subsection{Set Based Measures}

In this section we define three simple scores based on the set of overlapping arguments between the antecedent and consequent relation: DIRT, Cover and BInc.

\begin{defn}
For notational convenience, it is often useful to refer to the set of all arguments found to be instantiating a given argument. For a given relation $r$, we define this set to be:

$$\T_r \defeq \{r \mid r(t) \in \O\}$$
\end{defn}

To illustrate, in \sw{} we would have:

\begin{align*}
\T_\texttt{studies-at} &= \{\texttt{(Jane, Macquarie), (Sam, Macquarie)}\} \quad\text{whereas:}\\
\T_\texttt{tutors-at} &= \{\texttt{(Sam, Macquarie), (Jacob, Macquarie)}\}
\end{align*}

This representation along with standard set operators allow for a compact and expressible notation of key ideas. For example, the set of tuples observed to instantiate both $p$ and $q$ is given by $\T_p\cap\T_q$, and the set of tuples observed to instantiate either $p$ and $q$ can be expressed as $\T_p \cup \T_q$.

\begin{description}

\item[DIRT] DIRT \cite{Lin} is measure of overlap. For an implication $p\rightarrow q$, DIRT measures the weighted proportion of instantiating arguments that overlap between $p$ and $q$. The weighting occurs on a relation-tuple level -- we denote the weight of relation $p$ and argument $t$ as $w(p,t)$. In the original DIRT paper, the weight of a relation and argument was defined to be the pointwise mutual information (PMI) of $p$ and $t$\footnote{Where for two events $x,y$ belonging to discrete random variables $X,Y$, $PMI(x;y) \defeq \log\frac{\P{X=x,Y=y}}{\P{X=x}\P{Y=y}}$.}.

$$\text{DIRT}(p\rightarrow q) = \frac{\sum_{t \in \T_p\cap \T_q} w(p,t) + w(q,t)}
    {\sum_{t \in \T_p } w(p,t) + \sum_{t \in \T_q } w(q,t)}$$
    
DIRT ranges from from $0$ to $1$. The DIRT score for two relations $p,q$ is minimised when $T_p\cap\T_q = \emptyset$: that is, $p$ and $q$ are not instantiated by any of the same tuples. In contrast, DIRT is maximised when $\T_p \cap \T_q = \T_p \cup \T_q$, i.e. $p$ and $q$ are instantiated by the exact same set of arguments.

We note that due to commutativity in the set intersection and numeric addition operators, DIRT is a symmetric measure. That is, $\text{DIRT}(p\rightarrow q) = \text{DIRT}(q\rightarrow p)$ for all $p,q$. This is undesirable for the task of relational implication, as clearly implication rules can and often are asymmetric. One simple example is with with hypernymic implications -- $\texttt{tutors-at} \rightarrow \texttt{works-for}$ with high probability, but not vice versa. 

In order to illustrate how DIRT is calculated, we once more refer to \sw{}. Here, we set $w(r,t) = 1$ for all $r,t$ for simplicity. We find:

\begin{align*}
\text{DIRT}(\texttt{tutors-at} \leftrightarrow \texttt{works-for}) &= \frac{1}{2} \\
\end{align*}

Where we have used $\leftrightarrow$ to reinforce the symmetry of DIRT.
This score corresponds to the fact that \texttt{tutors-at} and \texttt{works-for} share half of the same argument tuples. However, this means we also conclude \texttt{works-for} $\rightarrow$ \texttt{tutors-at} with the same degree of belief.

Contrastingly, we find:

\begin{align*}
\text{DIRT}(\texttt{taught-by} \leftrightarrow \texttt{works-for}) &= \frac{0}{4} 
\end{align*}

Which can be seen from the fact that the two relations share no argument-tuples. 
 
\item[Cover] Cover \cite{Cover} is an alternative to DIRT that handles similarity asymmetrically. The Cover score of an implication $p\rightarrow q$ is defined as:

$$\text{Cover}(p\rightarrow q) = \frac{\sum_{t \in T_p\cap T_q} w(p,t)}
{\sum_{t \in T_p } w(p,t)}$$

Cover can be seen as the weighted proportion of the arguments of the antecedent relation that are shared by the consequent. Intuitively, this makes sense -- if $p \rightarrow q$, then whenever we see arguments instantiating $p$, we would expect them to also be compatible with $q$. However, we can still have arguments instantiating $q$ that aren't compatible with $p$ -- if \texttt{tutors-at} $\rightarrow$ \texttt{works-for}, then we expect all $t$ such that \texttt{tutors-at}(t) is true to also be compatible with \texttt{works-for}. On the other hand, there may be many non-tutor roles that fall under staff member and as such the converse is not true.

We also note that Cover resembles a measure of conditional probability. For example, if we ignore the weight terms, we find Cover$(p\rightarrow q)$ evaluates to:

\begin{align*}
    \sum_{t\in\mathcal{T}} \frac{\ind{t\in \mathcal{T}_p \cap \mathcal{T}_q}}
                            {\ind{t\in \mathcal{T}_p}}
\end{align*}

Where the numerator can be seen (roughly) as a measure of the joint probability of $p$ and $q$, and the denominator (roughly) as a measure of the probability of $p$. For two random variables $X,Y$ $\frac{\P{X,Y}}{\P{X}} = \P{X \mid Y}$ -- and so Cover in form resembles the conditional probability of $q$ knowing $p$ to be true. 

We again reference \sw{} for illustration. Setting all weights to one, we find that:

\begin{align*}
\text{Cover}(\texttt{tutors-at} \rightarrow \texttt{works-for}) &= 1 \\\\
\text{Cover}(\texttt{works-for} \rightarrow \texttt{tutors-at}) &= \frac{1}{2}
\end{align*}

Which is a much more reasonable representation of our belief. Wherever we have seen a tutor, they have also been a staff member. On the other hand, when we have observed a staff member they have also been a tutor only 50\% of the time. 

\item[Balanced Inclusion] BInc (Balanced Inclusion) \cite{BInc} is a simple aggregate score. BInc, for an implicaition $i\rightarrow j$ is defined to be:
    
$$\text{BInc}(i\rightarrow j) = \sqrt{\text{Lin}(i\rightarrow j) \cdot \text{Cover}(i\rightarrow j)}$$

That is, the geometric average of the DIRT and Cover scores for a candidate implication. It is unsupervised and highly simplistic. Nevertheless, it has been demonstrated to improve performance on several evaluation tasks when compared to the constituent scores \cite{BInc}.

\end{description}

\subsubsection{Feature Representations in Relational Implication}
\label{subsec:feature_rep}

We note that in its declarative paper, DIRT was calculated over a different set argument representation than the one we see here. Specifically, two DIRT scores were calculated -- one for the subject slot, and one for the object. The two measures were combined together by geometric average constituting the final score. Analogously, Cover was first calculated using a unary decomposition. A binary implication was decomposed into a number of unary implication rules, and the final score was the average of pairs of antecedent/consequent unary propositions.

To illustrate this, we generalise our scoring functions given above. Rather than calculating relevant quantities over the sets $\T_p, \T_q$, more general notions of $\mathcal{F}_p$ and $\mathcal{F}_q$ are specified as a model parameter. For example, setting $\mathcal{F}_p = \{s \mid p(s,o) \in \O \}$ and $\mathcal{F}_q = \{s \mid q(s,o) \in \O \}$ would correspond to the case where we simply count overlapping arguments in the subject-slot. We use this to define the feature representations described above.

\begin{defn}
Argument representations in DIRT, Cover and BINC:
\begin{description}
\item[i. Argument-tuple] A single score is calculated for the implication rule with $\mathcal{F}_p = \T_p$ and $\mathcal{F}_q = \T_q$ as above.
\item[ii. Slot independent] Two scores are calculated for the implication rule. The first corresponds to the subject slot and uses: 
\begin{align*}
\mathcal{F}_p &= \{s \mid p(s,o) \in \O \},\\ \mathcal{F}_q &= \{s \mid q(s,o) \in \O \}
\end{align*}
The second corresponds to the object slot and uses \begin{align*}
\mathcal{F}_p &= \{o \mid p(s,o) \in \O \},\\ \mathcal{F}_q &= \{o \mid q(s,o) \in \O \}
\end{align*}
The two scores are combined by geometric average.
\item[iii. Unary] In a unary feature representation, four scores are calculated corresponding to all pairs of antecedent/consequent subject/object slots. Specifically, we have:

\begin{align*}
\mathcal{F}_p &= \{s \mid p(s,o) \in \O \},\\ \mathcal{F}_q &= \{s \mid q(s,o) \in \O \} \\ \\ 
\mathcal{F}_p &= \{s \mid p(s,o) \in \O \},\\ \mathcal{F}_q &= \{o \mid q(s,o) \in \O \}\\ \\ 
\mathcal{F}_p &= \{o \mid p(s,o) \in \O \},\\ \mathcal{F}_q &= \{s \mid q(s,o) \in \O \}\\ \\ 
\mathcal{F}_p &= \{o \mid p(s,o) \in \O \},\\ \mathcal{F}_q &= \{o \mid q(s,o) \in \O \}
\end{align*}
Again the final score is taken to be the geometric average of the four values corresponding to the above feature representations.
\end{description}
\end{defn}

All three feature representations have been used to score DIRT, Cover and BInc \cite{berant}. However, where they have been used they have been in aggregate, in that there have been no results provided on the performance of each feature representation independently of the others. We provide the first such results in Chapter~\ref{ch:exp}.

\subsection{Entailment Graph}

The entailment graph is a large scale system for learning relational implication rules devised by Berant et al. \cite{berant}. Their results are based on: 1) a semi-supervised classifier of entailment and, 2) an integer linear programming (ILP) formulation of the problem that adjusts these local scores in an intelligent way. 

The former, which they denote the local classifier, first calculates a large number of existing measures of implication between all pairs of relations (including DIRT, Cover and BInc). They then train a logistic regression model on top of these representations, using them as features in an ensemble or meta classifier. This ensures that each score is weighted commensurate to its effectiveness, and provides a more principalled approach to aggregation than simply taking an average. 

They generate their training data in a semi-supervised manner from WordNet. For example, say two relations $p,q$ differ from one another by a single token $t_p \neq t_q$. If WordNet suggested that $t_q$ was a hyponym of $t_p$, then $p\rightarrow q$ would be included as a positive training example. On the other hand, if WordNet stated that $t_p$ and $t_q$ were antonyms, $p\rightarrow q$ would be included as a negative training example. 

The second component of their model is an ILP formulation which uses the assumption of transitivity in implication. That is, if $p\rightarrow q$  and $q\rightarrow s$; then $p\rightarrow s$. The authors construct an entailment graph, where nodes correspond to relations and directed edges correspond to the strength of relational implication as determined by the local score function. The goal then is to find the subset of edges that maximises the edge-weight sum. Notationally, $x_{pq}$ is a binary variable indicating if $p\rightarrow q$ is taken to be an implicational rule, and $w_{pq}$ is the local score for $p\rightarrow q$. The problem formulation becomes:

\begin{align*}
\max_x &\sum_{(p, q)\in \mathcal{R}\times\mathcal{R}} w_{pq}x_{pq}\\\\
s.t. \ &\forall \ (p,q,r) \in \mathcal{R}^3, & x_{pq} + x_{qr} - x_{pr} \leq 1\\
	&\forall \ (p,q) \in \R^2 & x_{ij} \in \left\{0,1\right\}
\end{align*}

Where the first constraint ensures transitivity and the second enforces that all edges either belong or do not belong to the final set of implication rules.

\subsection{Embedding Similarity}

To the best of our knowledge, there has only been one attempt at using dense representations of relations to score relational implication. Levy and Dagan \cite{ds:levy} use a link prediction model in order to learn dense representations for relations and arguments (described in more detail in Section~\ref{sec:embed} \emph{vide infra}). Their approach, referred to as relation embeddings in their work does not use the link prediction score function directly. Rather, they use the link prediction framework to learn a vector representation for all relations. For an implication $p\rightarrow q$ with corresponding vector representations $w_p, w_q$, the score of the implication is given as $\cos(w_p, w_q)$, where $\cos(\cdot)$ is the cosine similarity between the two relational vectors.

This score has several great aspects. The first is computation: having trained the dense representations (a significant task, but one that only has to be done once), scoring an implication requires a single dot product. In contrast, measures such as DIRT and Cover require summing over potentially a large number of tuples -- while the denominator of these scores can be precalculated, the numerator has to be done on a per implication basis. Furthermore, for an implication rule $p\rightarrow q$, the amount of elements to be summed is given by $|\T_p \cap \T_q|$, which could theoretically be as large as $|\T|$. Naturally, the entailment graph framework is no better in terms of efficiency -- not only do all measures of local entailment (including DIRT, Cover and BInc) have to be calculated, for all pairs of relations, the ILP component was demonstrated to be NP-Complete\cite{berant}.

The abstracted, embedded representation makes reasoning about the nature of the score more difficult. However, as argued in Sections~\ref{sec:link_predict} and \ref{sec:embed}, our dense representations of relations share a lot in common with the dense representations of words used in word2vec \cite{mikolov2013distributed} and other word embeddings models. This comparison offers some intuitive understanding of why the cosine approach may work: although, as with word embeddings, to the best of our knowledge there is no strong formal justification for their performance \cite{goldberg2014word2vec}. Word embeddings map the vectors for words which appear in a similar context, close together. So too do relational embeddings -- but context in this case refers to the arguments which instantiate them. Cosine distance then tells us how similar the relational vectors are, and by proxy, how similar the arguments that instantiate the two relations are.

Unfortunately, cosine similarity is naturally symmetric -- as such, we argue it cannot provide the sole measure of relational implication. Nevertheless, the benefits detailed above, especially in regards to performance are significant. 

%% file: ch_litreview/sec_emb.tex
\section{Link Prediction Models}
\label{sec:embed}

The novel method we propose in Chapter~\ref{ch:meth} uses estimators of the probability a proposition is true in order to calculate a score of relational implication. In one variant of the model (\text{ProbL}), we use link prediction models to compute these estimators. As such, in this section we outline and describe link prediction from both a general perspective as well as describing specifically which models we have chosen for Chapter~\ref{ch:meth} and why. 

Link prediction models assign a score to arbitrary propositions (page~\pageref{def:link_pred}) by learning dense, latent representations for relations and arguments. In doing so, it becomes possible to identify highly probable but unseen propositions and include them in a knowledge base.

For pedagogy, a comparison can be drawn to word embeddings models. This is particularly apt, as the broad approach to solving both of the problems is the same: generalised matrix factorisation. In fact, the first relational model we discuss in this section can be trained directly using Google's word2vec framework \cite{mikolov2013distributed}. Word embeddings learn representations for words and the contexts in which they appear. Pairs of words which frequently share the same context are mapped to a similar location in the embeddings space \cite{goldberg2014word2vec}. In the link prediction task, relations and arguments take the place of words and their contexts. In analogue to above, we should then expect pairs of relations which share similar instantiating arguments to be mapped together in the dense embedding space. However, this property emerges as a useful side effect in both word embeddings and link prediction models. The training procedure, detailed below, optimises for a simpler goal.

\subsection{Framework for Link Prediction Models}

In general, link prediction models learn dense representations for each argument, not an argument tuple. As such, for notational convenience in this section, we decompose the argument tuple into its constituent subject and object components, i.e. $t = (s,o)$.

Broadly speaking, link prediction models define a parametisation or embedding space for each relation and argument. We represent this set of possible transformations as $\Theta$. In the simplest case, $\Theta$ might be comprised of a vector in $\mathbb{R}^k$ for each relation and argument, where $k$ is a model parameter which controls model complexity. Link prediction models use this tensor of learned weights to provide a score to a given proposition $r(s,o)$ -- although the exact form of this scoring function is model dependent. In general, we denote a link prediction scoring function as follows:

$$\phi\left(r,s,o; \Theta\right)$$

We train $\Theta$ by defining a simple proxy task and using the results to iteratively adjust our learned representations. Specifically, we try to learn our embeddings such that true propositions receive a high score and false propositions receive a low score. The first caveat is that relational data is in general positive: $\O$ contains a set of (assumed true) propositions, but provides no information on propositions which are untrue. The approach adopted in both word embeddings models and all approaches to link prediction described below, is to generate a set of negative propositions through random sampling. This technique, known as negative sampling, works as follows: for each true proposition $r(s,o)$ taken from $\O$, a set of negative samples of the form $r(s',o')$ are taken, where $s',o'$ are sampled uniformly from $\A^2$.

Given a batch of true samples and a corresponding set of negative propositions, the task is to leverage and combine the score assigned to each proposition into an overarching loss function that indicates how well the model has performed on that batch. There are two main loss functions used in the literature for link prediction. The first is not specific to negative sampling methods, instead treating the task in a simple binary classification framework with one label for the true facts and another for the false facts. 

\begin{defn}
Binary cross entropy: for each proposition $r_i(s_i,o_i)$, set $y_i = 1$ if $i$ was a true proposition and $y_i = -1$ if $i$ was negatively sampled. For a batch of such propositions $B$, the binary cross entropy is defined as:

$$\mathcal{L}(B) =  \sum_{i \in B} \log\left(1 + e^{-y_i \cdot \phi\left(r_i,s_i,o_i; \Theta\right)}\right) $$
\end{defn}

This method treats all negatively sampled propositions independently of the true fact which generated them. Models trained using this loss function learn to generally assign true propositions to false propositions. In contrast, the other common loss function, pairwise loss, makes use of the particular paired structure generated by negative sampling. Here, the task is to assign a higher score to a true proposition than specifically the negatively sampled propositions generated thereby.

\begin{defn}
Pairwise loss: for each true proposition $r_i(s_i,s_i)$, let $n(i)$ indicate the set of negative samples generated from $r_i(s_i,s_i)$. Furthermore, let $B$ indicate a batch, and $B^+$ the true propositions within this batch. We have:

$$\mathcal{L}(B) = \sum_{i \in B^+} \sum_{r_i(s_i', o_i') \in n(i)}
            | \phi(r_i, s_i', o_i';\Theta) - \phi(r_i, s_i, o_i;\Theta) |$$

\end{defn}

The derivative of the loss function is defined with respect to $\Theta$, and this is used to perform gradient descent and update the embedding weights. Having sketched the requisite components, we restate them here briefly for clarity:

\begin{itemise}
\item Randomly initialise $\Theta$.
\item Generate a batch, $B$, of true and negatively sampled propositions.
\item Score each proposition in $B$ using $\phi$.
\item Use the results to calculate the batch loss $\mathcal{L}(B)$.
\item Update the set of learned weights using the matrix of partial derivatives with respect to $\mathcal{L}$. For a given learning rate parameter $\eta$, we use the rule\footnote{In practice, more sophisticated update rules are used.}:
$$\Theta \gets \Theta - \eta \frac{\partial \mathcal{L}(B)}{\partial \Theta}$$
\item Repeat all but the first step either for a fixed number of iterations or until the magnitude of changes in $\Theta$ drops below a threshhold.
\end{itemise}

There are a range of additional modelling decisions which are not tied to a specific scoring or loss function but nevertheless can have a noticeable impact on model performance. In particular, model regularisation, where sparsity of model weights is encouraged through penalising their magnitude, was shown to improve performance across the board (given suitable hyperparameter selection)\cite{complex}. Hyperparameter selection in general is important, as is the variant of gradient descent used.

\subsection{Model Criteria}

There are several axes over which to assess link prediction models. In particular, we argue that the current model training and evaluation procedures are not sufficient to differentiate models based solely on performance. Non model specific factors such as training framework, regularisation parameter, optimisation algorithm, and other hyperparameter tuning may confound existing results. For example, Trouillon et al. obtained substantially higher scores implementing several existing models (TransE, DistMult) when compared to the results of the respective original papers \cite{complex}. This was attributed to using an optimisation algorithm with an adaptive learning rate, and including a regularisation parameter. Furthermore, in preliminary results, Kadlec et al. demonstrate through extensive hyperparameter tuning DistMult can outperform all existing benchmarks on FB15k and WN18 \cite{kadlec2017knowledge}.

As such, the field's evaluatory and development pipeline may not be sufficiently mature to allow for a meaningful evaluation based simply on results. Therefore in addition to model performance, we define several theoretically and empirically desirable properties. 
\begin{description}
\item[Empirical performance] The first score criteria is simply how well the model performs on standard evaluation tasks. The most common measure of performance is filtered or unfiltered mean reciprocal rank. Similar to negative sampling, each test proposition $r(s,o)$ is ranked against all permutations of one argument slot, -- that is, $r(s',o)\cup r(s,o')$ where $s', o' \in \mathcal{A}^2$. The better the rank of a true proposition, the better the score. The mean rank is taken across true propositions, and the reciprocal of this value provides the final score. In the filtered variant of this task, corrupted propositions observed in the dataset are not ranked over.
\item[Assymetry] 
Our link predictions models should be capable of expressing propositions that are not commutative in the argument slots. That is, while \fact{tutors-at}{Sam}{Macquarie} is true, \fact{tutors-at}{Macquarie}{Sam} is not.  To this end, we define:

\begin{defn}
Symmetric model: a link prediction model $\phi$ is symmetric if it commutes in the argument slots. That is, if for all $r,s,o \in \R\times\A^2$, $\phi(r,s,o;\Theta) = \phi(r,o,s;\Theta)$. If this is not the case the model is non-symmetric.
\end{defn}

As we demonstrate, several high performing link prediction models are symmetric. This perhaps suggests that there is a lot of leverage to be gained simply by assigning high scores to triples that are argument compatible if not semantically meaningful. Perhaps this is an artefact of how link prediction models are evaluate -- MRR scores a proposition $r(s,o)$ by comparing it to all permutations $r(s',o)\cup r(s,o')$. This results in a large number (likely a majority) of truly nonsensical propositions. As such, perhaps MRR is rewarding those models which are biased towards these simpler cases.

\item[Memory constraints] Due to the scale of existing datasets (such as the Google Books relational dataset), training efficiently is important. Models which require quadratically large (in terms of the embedding dimension $k$) space or time to compute are undesirable. In particular, for efficient computation it is desirable to have all embeddings vectors stored in memory. This is of larger consideration when training on a GPU, as in general there is less GPU memory available than RAM.

\end{description}

\subsection{Link Prediction Models}

In this section, we specifically detail the models used in the experiment section of this thesis. We have chosen three models which have been demonstrated to perform well on task benchmarks -- TransE, DistMult and Complex \cite{complex,kadlec2017knowledge,transe}. In particular, Complex has the best known scores for MRR on WN18 and FB15K\footnote{See page~\pageref{tab:ds} for more information on these datasets.} in the literature. Additionally, we include the argument based matrix factorisation model of Riedel et al. \cite{riedel}. While this model has not been evaluated on any known relational implication benchmarks, it has been used to provide a score of relational implication and so we include it for comparison \cite{ds:levy}.

We use the Google Books corpus for training all of our models (page~\pageref{tab:ds}), which is several orders of magnitude larger than any other known set of relational data. As such, we constrain our model choice (and hence the models described in this section) to those with a linear memory footprint. While this means that certain high performance models are not considered, it may be the case that the additional training data afforded to us through an efficient choice of model will make up for the choice of a more simple model.

Note: we use $e_a$ to denote the embedded representation of $a\in\A$; $w_r$ likewise for $r\in\R$. A summary of the below descriptions can be found in table~\ref{tab:score_fn}.

\input{ch_litreview/model_table.tex}

\begin{description}
\item[MatrixFact] Riedel et al. present a family of link prediction models based on a simplistic matrix factorisation \cite{riedel}. In particular, we use make use of the `entity model', hereinafter known simply as the matrix factorisation or MatrixFact model in this thesis. The score function is defined to be: 

$$\phi(r,s,o; \Theta) = e_s^Tw_r +  e_o^Tw_r$$

In this variant, the score of a triple is decomposed into the sum of two argument-relation dot products. The score is clearly symmetric. However, it is efficient in terms of both memory and computational time -- requiring two dot-products to score a triple (or equivalently, one vector-addition operator and a single dot-product).

\item[DistMult] DistMult is a simple dot-sum score function based on a real valued vector decomposition. The score function is defined as: 

\begin{align*}
\phi(r,s,o;\Theta) &= e_s^T W_r e_o \\
                &= \sum_{i} e_{si}\cdot w_{ri}\cdot e_{oi}
    \end{align*}

Where $W_r$ is a diagonal matrix representing relation $r$ with diagonal elements $w_r$. The second equality follows from the fact that $W_r$ is diagonal, and we use this representation henceforth.

DistMult is symmetric due to the commutativity of scalar multiplication. It requires linear memory to store its argument and relation embeddings, and requires a single sum-product operation to score a proposition.

\item[Complex] Complex \cite{complex} was developed specifically with the goals of non-symmetry and efficiency in mind. They adopt a very similar vector decomposition structure as the DistMult model, and their scoring function looks very similar. However, by using complex instead of real value vectors, they can achieve a non-symmetric score while still maintaining time and space efficiency. The model is defined as:

$$\phi(r,s,o; \Theta) = \text{Re}\left(\sum_{i} e_{si}\cdot w_{ri}\cdot \bar{e}_{oi}\right)$$

Here $\text{Re}(\cdot)$ to indicate the real projection of a complex number or vector, ensuring the final link prediction score is still real valued.

The vector representation for an argument $e_a$ or relation $w_r$ are elements of $\mathbb{C}^k$ (where $k$ is the embedding size). Each element is comprised of a real and imaginary component, which for a complex number $z$ is denoted as $z = \text{Re}(z) + i\cdot\text{Im}(z)$.

The justification for this is as follows: While both DistMult and Complex are based on vector-matrix dot products, the real vector dot product is commutative and the complex vector dot product is not. By definition, $u\cdot v = u^T\bar{v}$, where $\bar{v}$ indicates the complex conjugate of $v$\footnote{The complex conjugate of a vector $z$ is defined as $\text{Re}(z) - i \text{Im}(z)$.}. As such, their model, based on a complex-matrix decomposition takes the conjugate of only the object vector, and is hence non-symmetric.

In terms of time and space complexity, Complex is worse than DistMult but only by a small constant factor. In terms of memory, in order to store a $k$ dimensional complex vector is is sufficient to store the real and imaginary components. This means that, in total, $2k$ values will need to be stored per relation and argument. Similarly, it was demonstrated \cite{complex} that taking the real projection of the complex value product-sum above can decomposed into four real valued product sums.

\item[TransE] TransE \cite{transe} has a different structural form compared to previous models. The score function is defined as:

$$\phi(r,s,o; \Theta) = -\|(e_s + w_r) - e_o\|_p $$

Where $\|\cdot\|_p$ indicates the p-norm. Typically either $p = 1$ (corresponding to the Manhattan norm) or $p=2$ (the euclidean norm). The original paper presents the right hand side without the `$-$' sign as a measure of dissimilarity -- a proposition is more likely to be true if the score is low. In order to make interpretation consistent, we have transformed the measure of distance into one of similarity by negating it.

The basic idea behind TransE is that the relationship between two arguments can be represented by a vector-translation: $e_s + w_r \approx e_o$. As such, the score based on this measure simply measures the magnitude (for a given p-norm) in difference between the left and right side.
\end{description}

%% file: ch_litreview/model_table.tex
\begin{sidewaystable}
\centering
\begin{tabular}{@{}lllll@{}}
\toprule
\textbf{Dataset} & $\boldsymbol{\phi}\mathbf{(r,s,o; \Theta)}$ & \shortstack[l]{\textbf{WN18}\\\textbf{MRR}} & \shortstack[l]{\textbf{FB15K}\\\textbf{MRR}} & \textbf{Symmetry} \\ \midrule
\shortstack[l]{Matrix\\Factorisation} & $e_s^Tw_r +,e_o^Tw_r$ & N/A & N/A & Symmetric \\[1.2ex]
TransE & $\|(e_s + w_r) - e_o\|_p $ & 0.454 & 0.380 & Non-symmetric \\[1.5ex]
DistMult & $\sum_{i} e_{si}\cdot w_{ri}\cdot e_{oi}$ & 0.822 & 0.654* & Symmetric \\[1.5ex]
Complex & $ \text{Re}\left(\sum_{i} e_{si}\cdot w_{ri}\cdot \bar{e}_{oi}\right)$ & \textbf{0.941} & \textbf{0.692} & Non-symmetric \\[1.5ex] \bottomrule
\end{tabular}
\caption{Summary of common link prediction models. Here $e_a$ denotes the embedding representation for argument $a$ and $w_r$ indicates the embedding representation for relation $r$. $\text{Re}(\cdot)$ indicates the real component of a complex vector and $\bar{e}$ indicates the conjugate of a complex vector $e$. MRR is a model's mean reciprocal rank (filtered) and WN18/FB15K are standard evaluatory datasets. *: better results have been recorded for DistMult \cite{kadlec2017knowledge}, however their work is still in a preliminary stage.}
\label{tab:score_fn}

\end{sidewaystable}

%% file: ch_litreview/sec_eval.tex
\section{Evaluation of Relational Implication}

There have been a range of approaches to evaluating the relational inference task. Most early work focused on downstream evaluation; where authors used the output of the task as component of a larger pipeline and evaluated based on downstream performance. More recently, there has been a concerted effort made to create a number of labelled datasets for the task. In this section, we describe the datasets used for the task in light of several key points of differentiation. We then describe the metrics that have been used to evaluate a model given a labelled dataset, as well as list of models, discussed in detail above, that have achieved state-of-the-art performance on one or more datasets.

\subsection{Labelled Datasets}\label{preannotate}

\begin{sidewaystable}[]
\centering
\begin{tabular}{@{}llllll@{}}
\toprule
\textbf{Dataset} & \textbf{Size} & \textbf{+ Rules} & \textbf{Annotation} & \textbf{Domain} & \textbf{Instantiation} \\[1.3ex] \midrule
Berant et al. (2011) & 39,012 & 3,427 & Expert & Limited & Typed \\[1.3ex]
Levy et al. (2014) & 20,336 & 2,220 & Expert & Limited & \shortstack[l]{Full instantiation,\\one argument shared} \\[1.3ex]
Zeichner et al. (2012) & 6,567 & 2,447 & Crowd & DIRT-biased &\shortstack[l]{Full instantiation,\\both arguments shared}\\[1.3ex]
Levy \& Dagan (2016) & 16,371 & 3,147 & Crowd & Representative &

\shortstack[l]{Subject typed, object\\instantiated}\\[1.3ex] \bottomrule
\end{tabular}
\caption{List of existing labelled relational implication rules. `+ Rules' refers to the number of positive instances per dataset.}
\label{tab:rel_dsets}
\end{sidewaystable}

To the best of our knowledge there are four sources of labelled relational implication data. Each contains a selection of antecedent and consequent relations, som along with a label indicating if there is an implication between the two.

We find that while the general form is the same, there are several key difference between the datasets. The first relates to argument instantiation. All datasets are annotated either by expert or crowd. It has been argued that annotation is difficult if there is no argument information provided -- for example, \texttt{kills} $\rightarrow$ \texttt{cures} is more ambiguous than \fact{kills}{Drug}{Disease} $\rightarrow$ \fact{cures}{Drug}{Disease}. As such, all datasets provide some argument information, but differ in precisely how this is done. One other key difference is in the coverage of these implication rules. Ideally the set of labelled examples should be sufficiently diverse and representative. This ensures that certain forms of implication aren't biased for or against, and provides a more accurate measure of model performance.

We refer to a dataset by its authors and year of publication. Summary information, including the size of the dataset and number of implication rules can be seen in table~\ref{tab:rel_dsets}. We also briefly describe the salient features of each dataset below before offering a more integrative critique.

\begin{description}
\item[Berant et al. (2011) \cite{Berant:2011te}] To the best of our knowledge this is the first labelled dataset of relational implication. It is an expert annotated corpus of moderate size -- with 3,427 positive implication rules. They instantiate relations with generic types, rather than specific arguments. That is one row might be of the form \fact{tutors-at}{PERSON}{ORGANISATION} $\rightarrow$ \fact{works-for}{PERSON}{ORGANISATION}. Additionally, while the taxonomy of possible types is quite large, they only provide annotations on ten of these. 
\item[Levy et al. (2014) \cite{Levy:2014wv}] This dataset was annotated by the authors and made public in 2014. It contains 2,220 positive implication rules and the relations are fully instantiated by arguments. However, only one argument has to be common across the antecedent and consequent. That is, an implication might be of the form \fact{tutors-at}{Sam}{Macquarie} $\rightarrow$ \fact{works-for}{Jacob}{Macquarie}. The domain is restricted to relations that occur in a medical context.
\item[Zeichner et al. (2012) \cite{Zeichner:2012wn}] This was the first dataset annotated through crowd workers. It is fully instantiated by arguements, and in contrast to above both subject and object are consistent between relations. Candidate implication rules were generated by first running DIRT, and then choosing the highest scoring relational rules to annotate. As such, the implications suffer from issues of recall -- rules which denote true implicational relations but which DIRT scores poorly will not be present in the dataset. 
\item[Levy and Dagan (2016) \cite{ds:levy}] This crowd annotated dataset was released in 2016. It directly addresses the issues of the Zeichner et al. (2012) set by defining a non-DIRT specific method for sampling candidate implication rules. Both arguments are shared between relations, however the subject is presented to the annotator in typed form. For example, \fact{tutors-at}{PERSON}{Macquarie} $\rightarrow$ \fact{works-for}{PERSON}{Macquarie}. \end{description}

We argue that the Levy and Dagan (2016) \cite{ds:levy} dataset is the best corpus in terms of form. It is the only one without a restrictive domain or recall bias. In terms of argument instantiation, the earlier Levy et al. (2014) dataset only shares one argument across antecedent and consequent. This confounds the tasks of relation and argument implication. For example \fact{tutors-at}{Sam}{Macquarie} $\rightarrow$ \fact{works-for}{Jacob}{Macquarie} is labelled non-implicational because of the arguments \texttt{Sam} and \texttt{Jacob}. In contrast, all other datasets, including Levy and Dagan (2016) share both arguments across relations.

While we argue that the form of this dataset is most suitable, the fact that it was crowd annotated (as opposed to expert) is less desirable. Relational implication can be a cognitively difficult task, and is confounded by modals (\texttt{lives-in} $\rightarrow$ \texttt{once-lived-in}) and generics (\texttt{is-spoken-in} $\rightarrow$ \texttt{is-the-language-of}). Crowd workers are highly motivated to work through annotation jobs quickly; so without the correct format, incentivisation and error checking it is not guaranteed that their output will be of sufficient quality. As a point of comparison, we note that Levy and Dagan spent \$375 USD for a dataset of 16,371 rows \cite{ds:levy} -- in contrast Zeicher et al. used \$1000 USD for a dataset of only 6,567 rows \cite{Zeichner:2012wn}. It is possible that these issues have no bearing on the quality of the final dataset. Regardless, we argue that this is a hypothesis that warrants further testing.

\subsection{Evaluation Protocol and Results}

Here we briefly describe the metrics used to evaluate system performance given a labelled dataset, as well as the relational implication models which scored highest. 

In general, the two most common approaches to evaluation were to provide an area under ROC curve (as in \cite{Berant:2011te}) score and/or a precision recall curve (as in \cite{Berant:2011te, ds:levy}). The former is a standard measure in statistics and indicates the performance of a model across all levels of false-positive rate. The area under ROC (hence AUROC) is also interpretable -- the AUROC for a model indicates indicates the expected likelihood that it differentiates a randomly sampled true implication from a randomly sampled false implication. Precision recall curve on the other hand provides a continuum of values visually, rather than a single score. In particular, the precision recall curve tells us the precision of a model for all levels of recall. 

In terms of model benchmarks, Berant et al. \cite{berant} evaluate Dirt, Cover and BInc as well as their entailment graph model. They found that the entailment graph provided the best results by a significant margin. On the Levy and Dagan (2016) dataset, the authors also provide a number of baselines and benchmarks. This included embedded-cosine models and the entailment-graph formulation as above. Consistent with previous results, the entailment graph provided the best results of all of the other benchmark models.

%% file: ch_method/chap_method.tex
\graphicspath{{.}}

\chapter{Method}\label{ch:meth}

Structuring data in a relational format is a useful way of representing our world knowledge. However, often our understanding of the world is uncertain -- and all uncertainties are not made equally. It is then reasonable to believe that relational data too could benefit from being viewed through a probabilistic lens. This is true especially where relational data is scraped directly from text. The facts derived therefrom can be noisy both as a result of imperfect scraping and parsing; or they could be a correct derivation of an inherently fake or biased fact.

Frequently in the literature, implication is treated as a formal statement of deduction with guaranteed logical consequent. However, when dealing with natural language this idealisation is compromised through polysemy and noise in the input data\footnote{For further discussion see page~\pageref{sec:intro_impl}}. For example, \fact{studies-at}{Person}{University} $\not\rightarrow$ \fact{student-of}{Person}{University} -- perhaps `Person' is temporarily un-enrolled, or perhaps they are simply a member of public using the library facilities. However, the former provides evidence for the latter and accounting for this could lead to better knowledge base representation.

We choose to represent relational implication through the lens of conditional probability. In deductive implication, $p\rightarrow q$ is logically equivalent to $\lnot(p \land \lnot q)$, i.e. we can't see both $p$ and $\lnot q$. Relaxing the consequent, we instead require that when we see $p$, we expect with high likelihood that $q$ is true. In a rough formulation, a measure of the relational implication of $p \rightarrow q$ can then be understood as $\P{\text{``$q$ is true''} \mid \text{``$p$ is true''}}$. In this section, we develop a formal probabilistic model to capture exactly this intuition.

\input{ch_method/sec_bayes}

\input{ch_method/sec_emp}

\input{ch_method/sec_link}

\section{Concluding Remarks}

Relational implication, especially over noisy data is best viewed probabilistically. This is largely the basis of existing measures, which provide a continuous score between zero and one indicating the degree of implication. In fact, some existing measures such as Dirt and Cover (page~\pageref{sec:scores}) resemble a measure of conditional probability. However, these scores of relational implication are still only heuristically motivated. In this section, inspired by previous work, we formalise a probabilistic representation of implication rule. We provide three variants: ProbE, ProbL and ProbEL. By evaluating the performance of each with respect to previous methods, we can address the research questions motivating this work. In particular, we can determine whether our probabilistic framework is useful in and of itself by comparing its performance to measures such as Cover and BInc: these measures are calculated over the same set of tuples, and both involve simple count statistics, link prediction models. On the other hand, by examining how ProbL performs in relation to ProbE, we can isolate the improvement our link prediction models provide overall.

%% file: ch_method/sec_bayes.tex
\section{A General Probabilistic Framework For Relational Implication}

We first argue the general form of our probabilistic framework. Relational implication can be interpreted as the likelihood of one relation given another. In order to capture this probabilistically, our model assigns a distinct random variable for each observed relation: we define $Z_r$ be a boolean valued random variable associated with the relation $r$. As we are concerned with relational implication where the antecedent and consequent share arguments (page~\pageref{sec:intro_impl}), we demonstrate that it is sufficient to model the set of argument-tuples in a single random variable. To this end we let $T$ (with domain the set of observed argument-tuples $\T$) represent our arguments\footnote{We note that for notational convenience the support of $T$ is a set of argument-tuples i.e. it is non-numeric. This is well defined, and is also referred to as a random element to differentiate from the more standard random variable with numeric support. It is also possible to simply define a mapping between the argument-tuples and the set of integers, and represent a tuple by its index.}.

We make some assumptions of conditional independence in order to ensure computability. In our framework, relations are conditionally independent given their instantiating tuple. We write:
 
\begin{align*}
    \P{T, Z_{1}, Z_{2},\dots,} = \P{T} \prod_{r \in \R} \P{Z_r | T} \numberthis \label{fn:joint}
\end{align*}

In this formulation, the probability that $t\in \T$ and $r\in \R$ are compatible is given by $\P{Z_r  = 1 \mid T = t}$. The probability of an implication rule $p \rightarrow q$ can be expressed in our model as $\P{Z_q = 1 \mid Z_p = 1}$. Our formulation is also capable of expressing more sophisticated statements of logical inference. For example, compositional implication of the form $p\land q \rightarrow r$ can be evaluated simply by calculating a score of $\P{Z_r = 1 \mid Z_p = 1, Z_q = 1}$. Implicational negation, where $p\rightarrow \lnot q$ is also trivially represented within this formulation -- we simply calculate $\P{Z_q = 0 \mid Z_p = 1}$. This is beyond the scope of this thesis, nevertheless it is a promising avenue for future work and hence we mention it in passing.

\begin{figure}[ht]
\centering
\includegraphics[width=8cm]{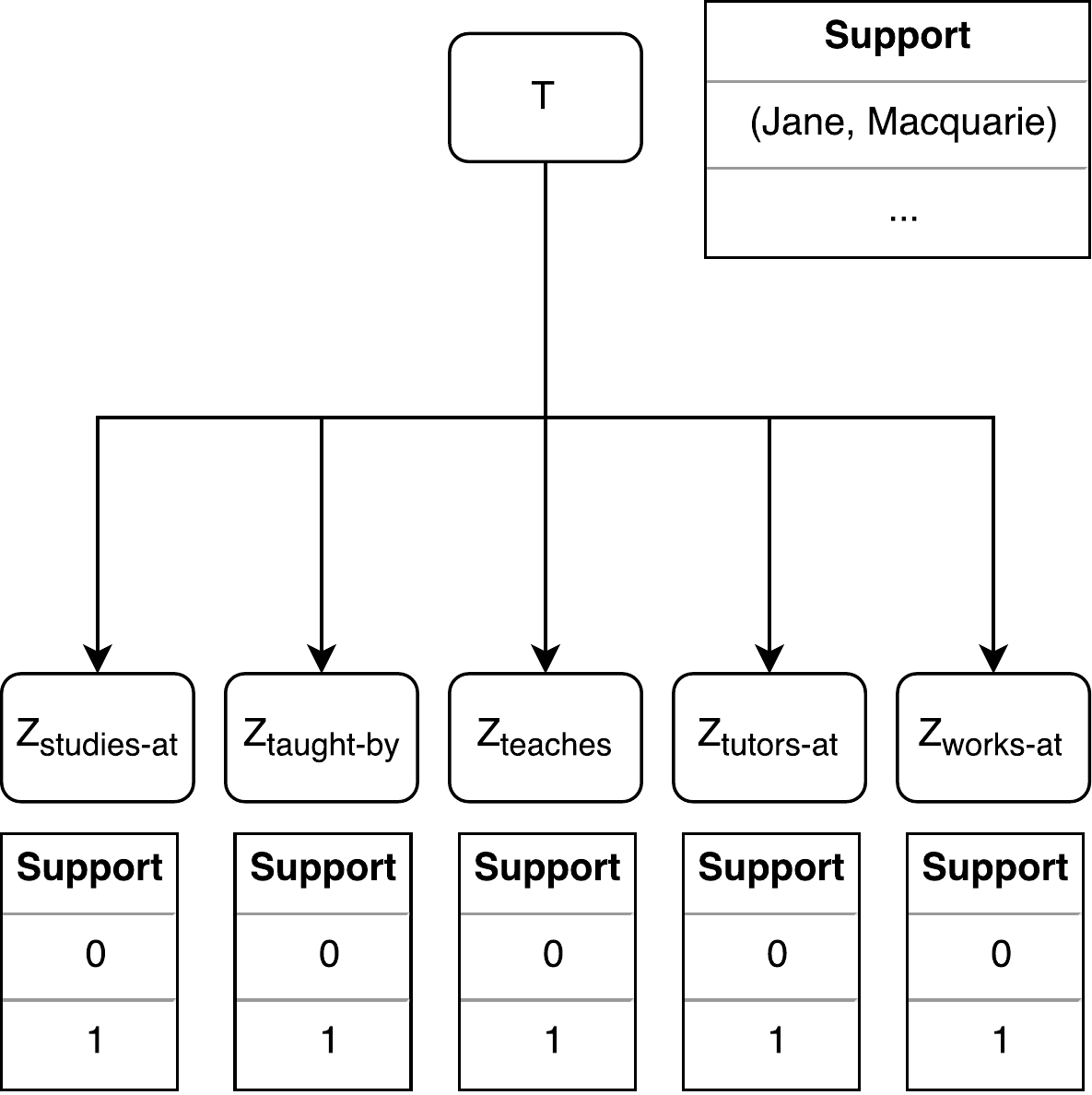}
\caption{Joint probability function on \sw{} dataset, represented as a Bayes net. A node represents a random variable and an edge denotes conditional dependence.}
\label{fig:sw_bayes}
\end{figure}

Consider \sw{}. We can represent the joint probability function for this dataset as a Bayes net, and have done so for illustration in figure~\ref{fig:sw_bayes}. The probability of \fact{studies-at}{Jane}{Macquarie} is given by $\P{Z_\texttt{studies-at} = 1 \mid T = \texttt{(Jane, Macquarie)}}$. Similarly, we would score the likelihood of \texttt{tutors-at} $\rightarrow$ \texttt{works-for} as: 

$$\P{Z_\texttt{tutors-at} = 1 \mid Z_\texttt{works-for} = 1}$$ We demonstrate how to represent queries of this form in terms of more elementary probabilities below.

Using our joint distribution function (\ref{fn:joint}), we can decompose our relation conditoinals as follows:

\begin{align*}
\P{Z_q \mid Z_p}&= \sum_{t \in \T} \P{Z_q, T = t \mid Z_p } \\
                    &=\frac{\P{Z_p, Z_q, T = t}}{\P{Z_p}}
                    \\ &= 
    \frac{\sum_{t \in \T} \P{T=t}\P{Z_p \mid T = t}\P{Z_q \mid T = t}}{\P{Z_p}}\numberthis \label{fn:cond}
\end{align*}

That is, to calculate the score of $\P{Z_p = 1 \mid Z_q = 1}$ it is sufficient to define the marginal distributions of our argument-tuples and relations; and the conditional distributions of a relation given a specific argument. As we demonstrate below, there are a number of simple and intuitive estimators for these probability functions. We use three, and denote the resulting models ProbE, ProbL and ProbEL.

%% file: ch_method/sec_emp.tex
\subsection{Empirical Measure of Implicational Probability}
\label{sec:empirical}

Our first approach at defining estimators for (\ref{fn:cond}) is to use the empirical distributions of our set of input propositions. 
Let $n$ denote the size of our input set; $n_{rt}$ the number of times proposition $(r,t)$ was observed; $n_{r\bullet}$ the number of times a proposition with relation $r$ was observed; and $n_{\bullet t}$ the number of times a proposition with tuple $t$ was observed.

We define the relevant marginal and conditional distributions as follows\footnote{These definitions are intuitive, however due to our probabilistic framework, where each relation has an associated random variable, their derivation is not immediately rigorous. This is achieved by treating all observed facts r(t) as observations of $Z_r = 1, T = t$ and $Z_{r'} = 0 \ \ \forall r' \neq r$}:

\begin{align*}
    \Ptilde{Z_r = 1} &\defeq \frac{n_{r\bullet}}{n} \\ \\
    \Ptilde{T = t} &\defeq \frac{n_{\bullet t}}{n} \\ \\
    \Ptilde{Z_r = 1 \mid T = t} &\defeq \frac{n_{rt}}{n_{}}
\end{align*}

These measures are easy to calculate and interpret. For pedagogy, we once again refer to \sw{}. We find:

\begin{align*}
\Ptilde{Z_\texttt{works-for} = 1} &= \frac{2}{7} \\
\Ptilde{T = \texttt{(Sam, Macquarie)}} &= \frac{2}{7} \\
\Ptilde{Z_\texttt{works-for} \mid T = \texttt{(Sam, Macquarie)}} &= \frac{1}{2}
\end{align*}

To calculate the conditional likelihood of the consequent given the antecedent, we substitute our estimators into (\ref{fn:cond}):

\begin{align*}
\Ptilde{Z_q = 1 \mid Z_p = 1} &= \frac{\sum_{t \in\T}
    \Ptilde{T=t}\Ptilde{Z_p = 1 \mid T = t}
    \Ptilde{Z_q = 1 \mid T = t}
}{\Ptilde{Z_p = 1}}\\
&= \frac{\sum_{t \in\T_p \cap\T_q}
    \Ptilde{T=t}\Ptilde{Z_p = 1 \mid T = t}\Ptilde{Z_q = 1 \mid T = t}
}{\Ptilde{Z_p = 1}} \\
\end{align*}

Where the second equality follows from the fact that the numerator is only non-zero for $t$ such that $n_{pt} > 0$ and $n_{qt} > 0$. We note that by using this empirical estimation of the underlying densities we in fact score over the same tuples as DIRT, Cover and BInc (page~\pageref{sec:scores}).

We define the score calculated using empirical estimators as:

\begin{align*}
\text{ProbE}(p\rightarrow q) \defeq \Ptilde{Z_q = 1 \mid Z_p = 1}\numberthis\label{fn:probe}
\end{align*}

%% file: ch_method/sec_link.tex
\subsection{Link Prediction Models as Estimators}

It is possible that there are unobserved but likely triples. To this end, we use link prediction models to provide an estimator for our conditional distributions. Our belief is that we can improve the score of relational implicatoin by using this additional data in our calculations.

Recall that a link prediction model was defined by a score function $\phi$ which provides a learned weight to any possible proposition (page~\pageref{sec:embed}). We define a family of models parametised by a link prediction function $\phi$\footnote{Any link prediction function defines a valid parametisation, however we the quality of the learned representation will impact the quality of the relational score estimated using the model.}. 

\begin{align*}
\Psubhat{\phi}{Z_r = 1 \mid T = t} &\defeq \sigma\left(\phi\left(r,t; \Theta \right)\right) \\ \\ 
\Psubhat{\phi}{T = t} &\defeq \Ptilde{T=t}
\end{align*}

Where $\sigma(x) = 1/(1+\exp(-x))$ denotes the logistic sigmoid function.

That is, the conditionals are scored based on the link-prediction function and the marginal density of the argument-tuple is taken to be the empirical distribution (as defined above). As our estimator for the conditional densities have changed, we must recalculate the marginal distribution of each relation:

\begin{align*}
\Psubhat{\phi}{Z_i = 1} 
&=  \sum_{t\in\mathcal{T}} \Psubhat{\phi}{Z_i = 1 \mid T = t}\Psubhat{\phi}{T = t}
\end{align*}

By substituting into (\ref{fn:cond}) we define:

\begin{align*}
    \text{ProbL}_\phi(p \rightarrow q) \defeq \Psubhat{\phi}{Z_j = 1 \mid Z_r = 1}\numberthis\label{fn:probl}
\end{align*}

\subsection{Efficient Link Prediction Model}

We note that to evaluate an implication rule through ProbL requires $O(\mathcal{T})$ time. This can be computationally expensive, especially as our set of argument-tuples can be quadratic in the size of our observed arguments. For large datasets or argument dense datasets, such a method may not be applicable. Additionally, any methods aimed at discovering new implication rules (by calculating the relational implication score across all possible tuple-combinations) is simply not feasible under this model even for moderate sized datasets.

In order to address this we propose a combination of ProbE and ProbL which we define as ProbEL. In this variant, we do not score over unobserved relations, but instead use the link-prediction model to provide a less noisy measure of conditional probability for our arguments. ProbEL$(p \rightarrow q)$ is calculated over the same set of tuples as ProbE (and hence, Dirt and Cover) -- $\T_p\cap \T_q$. However, as with ProbL, we use a link prediction model to estimate the conditional distributions. We represent this family of estimators by $\hat{P}_\phi^*$, and define:

\begin{align*}
\hat{P}_\phi^*\left(Z_r = 1 \mid T = t\right) &\defeq
\begin{cases}
\sigma\left(\phi\left(r,t; \Theta \right)\right) & \text{if $t\in\mathcal{T}_r$,}\\
0 &\text{else.}
\end{cases} \\ \\ 
\hat{P}_\phi^*\left(T = t\right) &\defeq \Ptilde{T = t}
\end{align*}

We calculate our estimator of the marginal density $\hat{P}_\phi^*\left(Z_r\right)$ in an analogous way to ProbL, with the caveat that we only have to sum over those tuples which for which our conditional is defined:

\begin{align*}
\hat{P}_\phi^*\left(Z_r = 1\right) &= \sum_{t\in\T_r} \hat{P}_\phi^*\left(Z_r = 1 \mid T = t\right)\hat{P}_\phi^*\left(T = t\right)
\end{align*}

And hence, using (\ref{fn:cond}) we define:

\begin{align*}
    \text{ProbEL}_\phi(p \rightarrow q) \defeq
    \hat{P}_\phi^*\left(Z_q = 1 \mid Z_p = 1\right)\numberthis\label{fn:probl}
\end{align*}

%% file: ch_exp/chap_exp.tex
\graphicspath{{.}}

\chapter{Experiments}\label{ch:exp}

Here we detail a series of experimental results. We have three broad goals in this section. The first is to verify our pipeline for model evaluation. We have argued that the form of the Levy and Dagan (2016) \cite{ds:levy} dataset is appropriate for evaluating relational implication models. However, due to the difficulty of the task and the nature of crowd annotation, we first evaluate the existing labels on a random sample of 150 implication rules. We find that the crowd labels are inadequate for implication rules that only apply in one direction (i.e. not statements of equivalence), and attribute this to the original task definition. As such, our next section in this chapter describes our reformulation of the crowd annotation framework. We argue why our variant is more appropriate to the task, set much more stringent requirements on crowd annotators and detail a thorough and ongoing quality assurance protocol. We use this to reannotate all positive implication rules in the Levy and Dagan (2016) dataset, and demonstrate that by doing so we increase expert-crowd annotator agreement on this subset from 53\% to 95\%.

The second is to evaluate a broad number of variants and parameter configurations for existing models of relational implication and link prediction. For set-based models of implication, such as Dirt, Cover and BInc, we examine the effect of different argument representations (see page~\ref{subsec:feature_rep}). In particular, we find that the tuple representation uniformly outperforms the other two, and that accounting for relations where the argument roles are reversed (such as \fact{reads}{X}{Y} $\rightarrow$ \fact{read-by}{Y}{X}) is particularly important for this dataset. We also use a wide variety of link prediction models and hyperparameter configurations as parametisations for Cosine, ProbL and ProbEL implication models. These were trained and developed in a high performance relational implication framework we developed as part of this thesis. We make this code publicly available\footnote{\url{github.com/xavi-ai}}.

Finally, we evaluate the performance of the novel methods developed in this thesis, and examine them in light of our research questions. We demonstrate that ProbE results in a better AUC and precision recall curve when compared to all previous work. Furthermore, with the correct link prediction parametisation, we demonstrate that ProbEL and ProbL improve upon the current state-of-the-art even further.

\input{ch_exp/sec_reannot}

\input{ch_exp/sec_exp}

\input{ch_exp/sec_res}

\section{Concluding Remarks}

In this section we have justified our experimental and methodological framework. We turn now to our research questions. Firstly, we have demonstrated that a formal probabilistic model is beneficial for the task. ProbE does not use link prediction models, instead relying on simple corpus statistics. Nevertheless, we have demonstrated that it outperforms all previous work, with an AUC of 0.7915 against the previous best result of 0.7812 (BInc). Secondly, we have demonstrated the efficacy of using link prediction models to enhance results on the relational implication task. Our ProbL model, which uses link prediction to score the degree of implication over all tuples receives the highest AUC of all approaches -- both on the full dataset and the directional subset. Finally, we have demonstrated that the existing evaluation dataset for relational implication is not of sufficient quality. We address this by proposing a new annotation schema, targeted specifically at addressing the errors of the original crowd labels. We evaluate the precision of our crowd labels before and after reannotation, and find that this improves from 53\% to 95\%.

%% file: ch_exp/sec_reannot.tex
\section{Dataset Evaluation and Reannotation}

In this section we first evaluate the quality of an existing labelled dataset of relational implication \cite{ds:levy}. We identify that there is a high error rate when considering implications that only apply in one direction (i.e. where $X \rightarrow Y$, $Y \not\rightarrow X$). In particular, implications of this form contribute to a high error in precision -- crowd workers very often incorrectly label these rules in the incorrect direction.

We then design a new crowd annotation schema for addressing this issue, and use the CrowdFlower platform to reannotate a portion of the Levy et al. (2016) dataset \cite{ds:levy}. As our error is with precision, we specifically reannotate all rows that were originally identified as positively implicational. We demonstrate that expert-crowd agreement on this subset increases from 53\% in the original formulation to 95\% in the new variant.

\input{ch_exp/subsec_ds_eval}

\input{ch_exp/subsec_ds_annot}

%% file: ch_exp/subsec_ds_eval.tex
\subsection{Evaluation of Levy et al. (2016) Dataset}
\label{evalDS}

After initial ad-hoc evaluation, we discovered frequent errors in precision for the crowd labelling scheme. We hypothesise that this was attributable to implication rules that occur only in one direction -- $\texttt{tutors-at} \rightarrow \texttt{works-for}$ but not vice versa. We describe our formal evaluation framework below.

In order to evaluate this claim, we uniformly sample 150 rows from the original dataset originally labelled as implicational. We then used three English-native tertiary educated annotators to reclassify this selection of 150 rows.

In order to mitigate annotator bias, we provide only the pair of relations -- not the original direction of implication. That is, if a row was of the form $p\rightarrow q$, our annotators only saw the two relations $p,q$ -- where the order that $p$ and $q$ was determined randomly. 

In the initial task, judgements were binary -- there either was or was not an implication relation. This is reasonable when formulating the task for crowd workers who operate over rows rapidly. However given the benefit of expert annotators we define an extended set of labels. This provides us with the ability to isolate and identify exactly which kinds of errors are being made.

For the implication involving relations $p,q$, we use the following four classes in categorisation:

\begin{description}
    \item[Strictly implies] Where $p$ implies $q$ but not vice-versa.
    \item[Strictly implied by] Where $q$ implies $p$ but not vice-versa. 
    \item[Equivalent to] Where $p$ implies $q$ and $q$ implies $p$.
    \item[Not implicational] Neither $p$ or $q$ implies the other.
\end{description}

After this annotation procedure we realign our labels with the original dataset. If an implication rule $p\rightarrow q$ was presented in the order $p,q$, the expert labels were unchanged. On the other hand, if the order of the relations were inverted (and provided as $q,p$), then a label of `strictly implies' became `strictly implied-by' and vice versa. Labels of equivalence or non-implication remain unchanged. 

As a specific example, say the rule $\texttt{tutors-at} \rightarrow \texttt{works-for}$ was present in the original dataset. Annotator A sees the tuple `$\texttt{tutors-at}, \texttt{works-for}$' and labels it as `strictly implies'. On the other hand, annotator B sees the relations in the reverse order -- `$\texttt{works-for}, \texttt{tutors-at}$' and as such labels it as `strictly implied by'. When realigning, we would leave annotator A's judgement as is, but change annotator B's judgement to `strictly implies'. 

After realigning the new labels we can compare directly to the results of the Levy et al. set. In particular, we aggregate our expert judgements by majority vote. We propose that for cases where there is no majority (i.e. each of the three annotators select a different label), the row simply be discarded. However, in our reannotation of the Levy et al. subset this did not occur. We provide the final aggregate judgements in table~\ref{tab:valid}, and a sample of rows for which our expert label differed from the crowd in table~\ref{tab:evaldsres}.

\begin{table}[ht]
\centering
\begin{tabular}{@{}lll@{}}
\toprule
\textbf{Label} & \textbf{Count} & \textbf{Percentage} \\ \midrule
Strictly implies & 42 & 28\% \\
Strictly implied by & 49 & 33\% \\
Equivalent to & 37 & 25\% \\
Not implicational & 22 & 15\% \\ \midrule
Total & 150 &  100\%\\ \bottomrule
\end{tabular}
\caption{Aggregated expert annotation on 150 implication rules identified by crowd workers in the Levy et al. (2016) dataset \cite{ds:levy}.}
\label{tab:valid}
\end{table}

\begin{table}[]
\centering
\resizebox{\textwidth}{!}{%
\begin{tabular}{@{}llll@{}}
\toprule
\textbf{Original Antecedent} &  & \textbf{Original Consequent} & \textbf{Reannotation} \\ \midrule
(Animal, Once Lived In,  America) & $\rightarrow$ & (Animal, Lives in, America) & $\leftarrow$ \\
(Britain, Possesses, Body of water) & $\rightarrow$ & (Body of water, Surrounds, Britain) & N/A \\
(Disease, Can present with, Fever) & $\rightarrow$ & (Disease, Causes, Fever) & $\leftarrow$ \\
(Hormone, Promotes, Sleep) & $\rightarrow$ & (Hormone, Induces, Sleep) & $\leftarrow$ \\
(Person, Never wrote, A symphony) & $\rightarrow$ & (Person, Composed, A symphony) & N/A \\ \bottomrule
\end{tabular}%
}
\caption{Example of rules identified as implicational in the dataset Levy et al. (2016), along with our expert reannotation. Here $\leftarrow$ indicates `strictly implied by' and N/A indicates `non implicational'.}
\label{tab:evaldsres}
\end{table}

Recall that we sampled from positive implication rules in the original dataset. As such, if our expert annotators label `strictly implies' or `equivalence', this corresponds to agreement with the crowd workers. However, we find this only occurs $53\%$ of the time. By far the largest source of error in the original dataset came from the `strictly implied by' labels. We found that 49 of the 150 rows fell into this category. By comparison, only 22 of the rows were identified as `not implicational'. As such, we argue that the original crowd workers frequently identified when an implication rule existed between two relations, but were less capable of determining the direction of that rule. We address this issue directly in our reformulation of the task described below.

%% file: ch_exp/subsec_ds_annot.tex
\subsection{Re-Annotation of Levy et al. (2016) Dataset}

Our input are candidate implication rules given by an antecedent and consequent relation; and arguments instantiating those relations. The set of arguments between relations is the same, however the order may be reversed. For example, in the rule  \fact{read}{X}{Y} $\rightarrow$ \fact{read-by}{Y}{X} -- the argument set ($X,Y$) is the same, however the order has been reversed. 

Our reannotation process is designed to address the previous issues of mislabelled directional implications. We take great care to distil the task of relational implication to its simplest form, and to prime the annotators to the importance of implication direction. 

For each rule $p \rightarrow q$ in the original dataset, we generate two annotation tasks. These correspond to the original direction of implication ($p\rightarrow q$), and the reversed direction of implication ($q\rightarrow p$). In order to eliminate a source of bias, the order in which these two rows are presented is randomised. However, to frame and focus our annotators on the importance of implication direction, these two rules are presented sequentially as a single task. That is, a crowd worker will see first an implication in one direction, and be asked to judge whether it is valid; and then the same implication rule in the other direction.

For example, if $\texttt{tutors-at} \rightarrow \texttt{works-for}$ was an original rule, our crowd workers would be asked to perform judgment on whether $\texttt{tutors-at} \rightarrow \texttt{works-for}$ as well as whether $\texttt{tutors-at} \rightarrow \texttt{tutors-at}$. We make it explicit in the task description the importance of directionality, and include several examples of uni-directional implication.

We also alter the way that a single judgement of relational implication is presented to the annotator. Levy et al. present multiple consequent relations for a single given antecedent relation. We instead simply ask the workers to examine a single antecedent and consequent relation. We also present the relational propositions in natural language, allowing for easier interpretation. In particular, for an implication rule $p\rightarrow q$, the natural language formulation is:

\begin{center}
If you know that $p_\mathtt{subject}\ \  p\ \  p_\mathtt{object}$,

Can you tell that $q_\mathtt{subject}\ \  q\ \  q_\mathtt{object}$?
\end{center}

Our earlier example would be structured as:

\begin{center}
If you know that $\texttt{PERSON}\ \  \texttt{tutors-at}\ \  \mathtt{UNIVERSITY}$,

Can you tell that $\texttt{PERSON}\ \  \texttt{works-for}\ \  \mathtt{UNIVERSITY}$?
\end{center}

\begin{figure}[ht]
\centering
\includegraphics[width=0.8\linewidth]{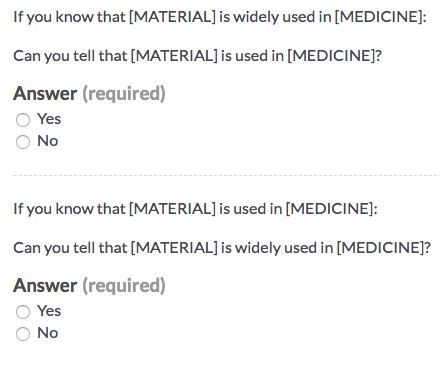}
\caption{An example of a judgement given to our crowd workers.}
\label{fig:crowdfl}
\end{figure}

We also present a screenshot from the crowd formulation task in figure~\ref{fig:crowdfl}.

Relational implication can be a difficult task and requires a strong level of literacy. As such, we endeavour to develop a robust and strict quality assurance protocol, whereby we accept a longer annotation period and higher annotation costs as a cost of more accurate predictions.

First, we restrict our task on CrowdFlower to `level three' annotators. The levels are a grade of annotator experience and performance, with three being the highest band representing less than 10\% of all annotators. We also select over one hundred test questions in order to represent a wide range of the forms in which relational implication can occur. The ground truth for these labels were determined by three expert annotator, and in order to ensure the task was fair we only included those implications which our expert annotators agreed on unanimously.

Before being eligible for the task, a worker is given a page of fifteen test questions comprised of two implicational judgements each. Annotators must score over 90\% accuracy on this set, when compared to the expert labels. Furthermore, each subsequent page of judgements presented to eligible workers contains one test question, and if at any point a worker's accuracy drops below the set threshold they are no longer eligible for the task.

Each judgement was annotated by at least four crowd workers, with additional judgements added dynamically for rows with low levels of annotator agreement. We provide annotation on 3,148 rows corresponding to 6,296 judgements of relational implication. The job cost \$201 USD in total, and was completed in just under a fortnight. 

In order to validate our results, we performed a slightly modified variant of the evaluatory procedure detailed in section~\ref{evalDS} with three expert annotators over 150 rows. In particular, because we had judgements by the crowd workers in both directions of implication, each row constituted two judgements and the mapping from the four label scheme differed slightly. For example, if the crowd indicated that $p\rightarrow q$ and that $q \rightarrow p$, then an expert label of `equivalent' would indicate two agreements of judgement. If the expert instead labelled the pair with a `strictly implies' rule, this would correspond to one agreement and one disagreement. Of the 150 rows and 300 constituent judgements, our expert annotators agreed with 285 or 95\%. 

We provide two datasets as a results of this process:

\begin{itemise}
\item Full dataset: As argued above, we are only required to reannotate the rules identified as implicational in the original dataset. As such, our final dataset is the union of all reannotated rules with the set of relations labelled as non-implicational in the original dataset.
\item Directional dataset: due to our enriched annotation procedure, we know for all reannotated implications whether implication occurs in one direction, both directions or if there is no implication. We use this to define a subset of the full dataset, comprised of all pairs of relations with an implication rule in strictly one direction (i.e $p\rightarrow q$, $q \not\rightarrow p$). For a given system, the task on this dataset is to indicate which direction is more likely for each row. This is a much more difficult task than general relational implication, because argument compatibility is not a sufficient measure here.
\end{itemise}

%% file: ch_exp/sec_exp.tex
\section{Experimental Framework}

In this section we describe our experimental framework, evaluatory metrics, model definitions and training scheme for link prediction. 

\begin{description}
\item[Datasets and evaluation] We provide results on the full and directional datasets described above. For the former, we provide both area under ROC curve (AUC) and precision-recall curve as is standard in the literature \cite{ds:levy, Berant:2011te}. For the directional dataset, the task formulation is slightly different. We treat it as a classification task, where models have to correctly identify the direction of implication. We evaluate performance based on accuracy i.e. the proportion of rules assigned the correct direction. On this set, models which are symmetric will score no better than chance.
\item[Models] We include all models discussed in Chapter~\ref{ch:lit}, as well as the novel methods defined in this thesis.

\begin{itemise}
\item DIRT, Cover, BINC: We implement three variants of each of these models, corresponding to the different feature representations identified in page~\pageref{subsec:feature_rep} (argument-tuple, slot independent and unary).
\item Cosine*: the cosine similarity calculated on embedded representations of the antecedent and consequent relations.
\item Entailment graph: We use Berant's publicly available entailment graph rules \cite{Berant:2011te}. As these are given as binary outcomes, they represent a single point in the precision-recall curve. This means that we cannot calculate AUC for this model.
\item ProbE: Our empirical model of probabilistic implication.
\item ProbL*: Our full link-prediction model of probabilistic implication.
\item ProbEL*: Our link-prediction model of probabilistic implication, calculated over tuples found to instantiate both antecedent and consequent relation.
\end{itemise}
*: Indicates that a model is parametised by a link prediction score. For all such models, we evaluate using MatrixFact \cite{riedel}, DistMult \cite{distmult}, Complex \cite{complex} and TransE \cite{transe}.

\item[Link prediction training] In order to assess the performance of a range of different relational embeddings model, design and develop a high-performance platform built on top of the Keras\footnote{\url{keras.io}} package, and we make our code publicly available.

Batch generation and negative example generation occurs in parallel on a system's CPUs and all prediction and back propagation occurs on the GPU. For our experiments we use a system with a single Nvidia Pascal graphics card and 40GB of RAM. 

For all models, we found best results using the Adam optimiser \cite{kingma2014adam} with Nesterov momentum \cite{dozat2016incorporating} and an initial learning rate of \texttt{0.001}. We dynamically anneal our learning rate and use a validation set to determine model termination validation.

We train over the Google Books relational corpus (page~\ref{tab:ds}). Due to the massive scale of this corpus, we restrict our argument and relation embeddings to vectors in $\mathbb{R}^{200}$. The exception to this is with Complex, which requires twice the amount of memory in order to store real and imaginary components. As such, the embeddings for this model are in $\mathbb{C}^{100}$.

\item[Implication with argument reversal] We find 6,400 implications in our dataset are of the form where the order of the arguments in the implied relation is reversed when compared to the implying relation -- for example, \fact{read}{person}{book}  $\rightarrow$  \fact{read-by}{book}{person}. As this equates to around 40\% of our implication rules, we believe it is important to address them -- and our empirical results verify this.

Following a similar approach to Berant et al. \cite{berant}, we define a new relation $r_{rev}$ for all $r\in \mathcal{R}$. Whenever we observe an input fact \fact{r}{s}{o}, we also treat it as an observation of \fact{$\text{r}_{\text{rev}}$}{o}{s}. In this way, an implication $p\rightarrow q$ of this form will be scored by $\P{Z_{q_{rev}} \mid Z_p}$. Similarly, for Dirt, Cover and BInc, we simply count over the overlapping arguments when considering the reversed relation (where applicable).

For our link prediction models, we simply permute the order of the arguments -- that is, $\phi(r_{\text{rev}},s,o;\Theta) = \phi(r,o,s;\Theta)$. Models with commutative argument slots will provide the same score to either order. However, it is worth noting that considering reversed implication rules will still effect empirical-link models, as the tuples over which scores are calculated will differ in the reversed representation.
\end{description}

Here we provide a brief justification for our evaluatory framework. As argued in Chapter\ref{ch:lit}, we believe the Levy et al. (2016) dataset to be the best form for the task. However, the annotation precision of their crowd workers is demonstrably insufficient. As such, we choose to use our reannotated corpus with higher precision. Our model selection includes, to the best of our knowledge, all previous relational implication models which have achieved high scores on the task. Similarly, for the link prediction methods parametising our ProbL and ProbEL methods, we evaluate a range of approaches.

%% file: ch_exp/sec_res.tex
\section{Results}

\input{ch_exp/small_tabs}

\begin{description}
\item[Feature representations] We compare the different feature representations used in Cover, DIRT and BInc (table~\ref{tab:res_feat}). We find that using an argument tuple based feature representation performs best across all scores. Unary feature representations outperformed the slot independent model, which is consistent with results shown by the authors of BInc \cite{BInc}. To the best of our knowledge, only Berant et al. have previously used an argument-tuple representation of features \cite{berant}. This is perhaps attributable to dataset size and sparsity issues -- there are quadratically more pairs of arguments than singletons. Without a dataset of sufficient size, many of the possible implication rules may be scored over too few arguments to get a meaningful result. All further results for DIRT, Cover and BInc are calculated using a tuple feature representation.

\item[Argument permutation] We also demonstrate the importance of correctly handling implicational rules where the argument slots are reversed (e.g. \fact{read}{X}{Y} $\rightarrow$ \fact{read-by}{Y}{X}) (table~\ref{tab:res_dir}). There is a consistent substantial improvement in score when accounting for order sensitivity. As identified above, there are 6,400 implication rules that follow this form. This represents over 35\% percent of the total rules in our dataset, as such the impact of handling them appropriately is unsurprising. Other evaluation sets which exclude such implications, or do not have as many reversed argument implications not be impacted to the same extent.

This also offers a partial explanation as to why Cosine similarity and other approaches which are not amenable to an argument-reversed form perform poorly. For Cosine, it may be useful in the future to alter the link prediction training in order to allow for reversed implication rules. That is, for all \fact{r}{s}{o} in the input dataset, we add an additional training instance \fact{$\text{r}_{\text{rev}}$}{o}{s}. This would allow us to use cosine similarity on reversed argument implications, and theoretically substantially improve the performance on this dataset. However, while doubling the size of the input is not prohibitive, doubling the number of relations may well be.

\item[Final results] We present the final results of our models. As above, all set based approaches use tuple-representations and where a model can be altered to account for argument permutation we have done so. Table~\ref{tab:final_res} contains all results on the full and directional datasets. Figures~\ref{fig:prc} and \ref{fig:roc} show the precision-recall and ROC curves for the best scoring model of each type.

We note several key findings:

\begin{itemize}
\item The three novel methods we have developed in this thesis outperform all previous work. On the full dataset, our best scores for ProbE, ProbEL and ProbL were 0.7915, 0.7944 and 0.8143 respectively. In comparison, the highest scoring existing method was BInc with an AUC of 0.7812. This was also consistent across the directional dataset, where our worst performing novel method ProbEL still outperformed the best previous work.
\item The directional dataset indicates that despite our positive results the task is still far from solved. While an AUC of 0.8143 on the full dataset indicates that it is possible to between true and false implication rules with reasonable confidence, at best our accuracy on the directional dataset was 0.6792. In future, perhaps further model development should be aimed at differentiating the direction of implication.
\item We find that in general, ProbL was our best technique for relational implication. This is unsurprising when considering that ProbEL is also a powerful model in its own right: ProbEL and ProbL are similar models, but the latter is calculated over more tuples. However, the additional performance of ProbL comes at the expense of extra computational work.
\item In terms of link prediction models, Complex performed best parametising both ProbEL and ProbL for the full and directional dataset. This is consistent with results described in the Complex paper \cite{complex}.

The only case where another model outperformed Complex was MatrixFact parametising Cosine. In this variant, link prediction scores are not used directly. As such, more powerful and expressive models may not be necessary and this complexity could come at the expense of a clear representation of vector proximity (which Cosine is based upon). One further note: if we observe the performance of TransE on this model variant we see that TransE scores particularly poorly -- worst of all recorded results. This perhaps indicates that the structure of TransE does not result in Cosine distance being a meaningful measure. Perhaps this is because Cosine similarity is calculated by taking a normalised dot-product: while MatrixFact, DistMult and Complex all use variants of a dot-product in their score function, TransE uses vector addition instead.
\item While we only have partial results, the entailment graph does not seem to perform as well as was expected. We see in figure~\ref{fig:prc} that all other models have a higher precision value at the recall value of the entailment graph rules. However, we attribute this to the provided form of our entailment graph rather than any deficiencies in the model itself. We are only given a predefined set of rules, trained on a separate corpus -- in particular, a corpus several orders of magnitude smaller than the Google Books relational dataset used to train our other models. In order to compare directly, we would need to fully reimplement the entailment graph formulation. However not only is that beyond the scope of this body of work, it is also not clear that the model would be capable of scaling up to the required corpus size.

\end{itemize}

\input{ch_exp/main_img}

\input{ch_exp/main_results}

\end{description}

%% file: ch_exp/small_tabs.tex
\begin{table}[]
\centering
\begin{tabular}{cc}

\begin{minipage}{.5\linewidth}
\begin{tabular}{@{}lrrr@{}}
    \toprule
    Model & Tuple & Slot idp. &  Unary \\ \midrule
    DIRT    & \textbf{0.6921} & 0.6845 & 0.6906 \\
    Cover   & \textbf{0.7170} & 0.7129 & 0.7107 \\
    BInc    & \textbf{0.7260} & 0.7122 & 0.7188 \\ \bottomrule
    \end{tabular}
    \caption{Performance of different argument representations for DIRT, Cover and BInc. `Slot idp' indicates slot independent. The best representation for each model has been bolded.}
    \label{tab:res_feat}
\end{minipage} & 

\begin{minipage}{.5\linewidth}
    \begin{tabular}{@{}lrr@{}}
    \toprule
    Model  & \shortstack{Order\\insensitive} & \shortstack{Order\\Sensitive} \\ \midrule
    DIRT & 0.6921 & 0.7606 \\
    Cover & 0.7170 & 0.7805 \\
    BInc & 0.7260 & 0.7812             \\ 
    ProbE & \textbf{0.7271} & \textbf{0.7915}            \\\bottomrule
    \end{tabular}
    \caption{Importance of accounting for implications where the order of the instantiating arguments is reversed. The best performing model with and without accounting for order is bolded.}
    \label{tab:res_dir}
\end{minipage}
\end{tabular}

\end{table}

%% file: ch_exp/main_img.tex
\begin{figure}[ht]
\centering
\includegraphics[width=\linewidth]{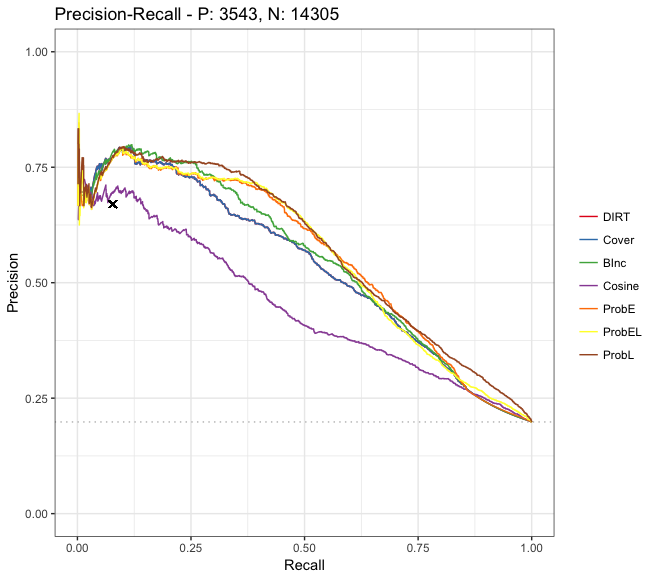}
\caption{Precision-recall curve of our best performing model variants. X: indicates the precision-recall values corresponding to the Berant et al. (2011) ruleset.}
\label{fig:prc}
\end{figure}

\begin{figure}[ht]
\centering
\includegraphics[width=\linewidth]{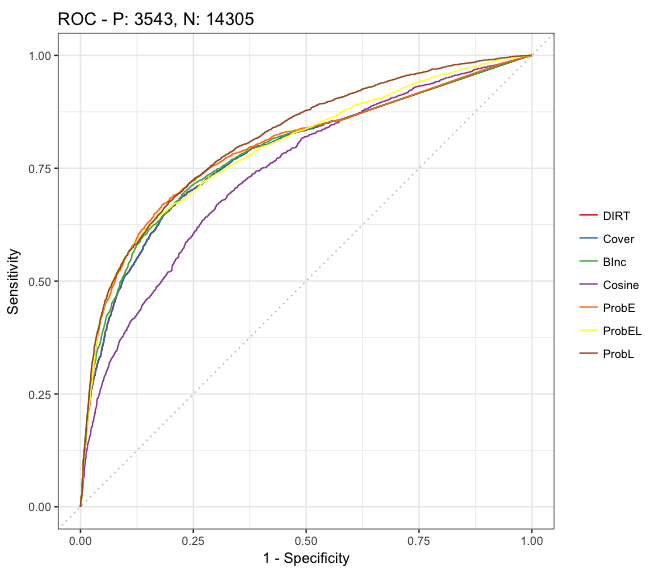}
\caption{ROC curve of our best performing model variants.}
\label{fig:roc}
\end{figure}

%% file: ch_exp/main_results.tex
\begin{table}[]
\centering
\begin{tabular}{@{}llrr@{}}
\toprule
& \textbf{Model}  & \textbf{Full (AUC)} & \textbf{Directional (ACC)} \\ \midrule
Set-based & DIRT & 0.7606 & -- \\
 & Cover  & 0.7805  & 0.6547 \\
 & BInc   & 0.7812 & 0.6547 \\
 &  &  &  \\
Cosine & MatrixFact & \emph{0.7413} & -- \\
 & DistMult & 0.7353 & --  \\
 & Complex & 0.7151 & -- \\
 & TransE & 0.6782 & -- \\
 &  &  &  \\
ProbE & ProbE & 0.7915  & 0.6606 \\
 &  &  &  \\
ProbEL & MatrixFact & 0.7633 & 0.6201 \\
 & DistMult & 0.7883 & 0.6149 \\
 & Complex & \emph{0.7944} & \emph{0.6581} \\
 & TransE & 0.7501 & 0.6498 \\
 &  &  &  \\
ProbL & MatrixFact & 0.7852 & -- \\
 & DistMult & 0.8044 & -- \\
 & Complex & \emph{\textbf{0.8143}}  & \emph{\textbf{0.6792}} \\
 & TransE & 0.7800 & 0.6372 \\ \bottomrule
\end{tabular}
\caption{Results of all models on full and directional dataset. AUC indicates area under curve, ACC indicates accuracy. `--' in the directional column indicates that the model is symmetric and as such performs no better than chance. The best link prediction model for each approach has been emphasised, and the best overall model for each task has been bolded.}
\label{tab:final_res}
\end{table}

%% file: ch_conclusion/chap_conclusion.tex
\chapter{Conclusion}
\label{ch:con}

This thesis has been concerned with relational data, in particular with the task of relational implication. From both a theoretical perspective, and from examining current methods of solving the problem, we have taken the position that this task is best viewed in a probabilistic context. However, while current approaches resemble probabilistic statements in form, they are not defined so explicitly. Doing so allows us to make clear what assumptions are underlying our work and provides a clear interpretable framework for model development and parametisation. Empirically, we demonstrate that the models we have developed herein using this framework outperform all previous work. This was attributable to both our probabilistic framework, and the inclusion of link prediction scores. As part of the evaluation process, we also demonstrated that current evaluatory datasets are not of sufficient quality. In order to address this, we developed a new targeted reannotation schema and made public the resulting dataset. 

There are several immediate avenues for future research. Our probabilistic model of relational data provides a broad and extensible platform for future work. It may be useful to use the model to evaluate compositional implication -- rules of the form $p\land q \rightarrow r$ (consider, for example, \texttt{student} $\land$ \texttt{studying-maths} $\rightarrow$ \texttt{maths-student}). This is possible within our model by simply calculating $\P{Z_{r} = 1\mid Z_{p} = 1, Z_{q} = 1}$. It would also be possible to extract inference rules of the form $p\rightarrow \lnot q$ -- simply calculate $\P{Z_q = 0 \mid Z_p = 1}$. Even more possibilities arise if we extend the model. For example, if we collected temporal information in addition to our propositions (which could be achieved by scraping propositions from news articles), we could develop a more thorough model of implication. For example, if $p$ was a prerequisite of $q$ then we would always expect $q \rightarrow p$ and also that $p < q$ (where $<$ indicates the left hand argument precedes the right). 

Relational data represents world knowledge and knowledge is often best represented probabilistically. This is the guiding principle behind link prediction models. On the other hand, relational implication has never formally been viewed within this framework. We have shown in this work that it is useful to do so in and of itself. Furthermore, we demonstrate that the probabilistic representation of link prediction models can be used to further improve our probabilistic representation of implication.

%% file: ch_appendix/chap_app.tex
\chapter{Appendix: \textsc{Smallworld}}
\label{chap:sml}

\input{smallworld.tex}

%% file: smallworld.tex
\begin{table}
\centering
\begin{tabular}{@{}lll@{}}
\toprule
Relation            & Subject        & Object             \\ \midrule
\texttt{studies-at} & \texttt{Jane}  & \texttt{Macquarie} \\
\texttt{studies-at} & \texttt{Sam}   & \texttt{Macquarie} \\
\texttt{taught-by}  & \texttt{Emily} & \texttt{Sam}       \\
\texttt{teaches}    & \texttt{Sam} & \texttt{Emily}       \\
\texttt{tutors-at}  & \texttt{Sam}   & \texttt{Macquarie} \\
\texttt{works-for}  & \texttt{Jacob}  & \texttt{Macquarie} \\ 
\texttt{works-for}  & \texttt{Sam}  & \texttt{Macquarie} \\ \bottomrule
\end{tabular}
\caption{\textsc{SmallWorld}: an illustrative set of relational data we refer to frequently in this thesis.}
\label{tab:smallworld}
\end{table}

%% file: listofsymbols.tex
\chapter{List of Symbols}


The following list is neither exhaustive nor exclusive, but may be helpful.
\begin{list}{}{%
\setlength{\labelwidth}{24mm}
\setlength{\leftmargin}{35mm}}
    \item $\mathcal{R}$: the set of all relations.
    \item $\mathcal{A}$: the set of all arguments.
    \item $\mathcal{T}$: the set of all observed argument (subject, object) pairs.
    \item $\mathcal{T}_r$: the set of all argument tuples that instantiate a given relation $r$, i.e. $\left\{t \mid (r,t) \in \mathcal{O}\right\}$.
    \item $\mathcal{O}$: the multiset of all observed facts, where each fact is comprised of a relation and an argument tuple.
    \item $\mathcal{O}_{r \bullet}$: the multiset of all observed facts with relation $r$.
    \item $\mathcal{O}_{\bullet t}$ the multiset of all observed facts with argument tuple $t$.
\end{list}

%% file: ms.bbl
\begin{thebibliography}{10}
\expandafter\ifx\csname url\endcsname\relax
  \def\url#1{\texttt{#1}}\fi
\expandafter\ifx\csname urlprefix\endcsname\relax\def\urlprefix{URL }\fi
\providecommand{\eprint}[2][]{\url{#2}}

\bibitem{kadlec2017knowledge}
R.~Kadlec, O.~Bajgar, and J.~Kleindienst.
\newblock \emph{Knowledge base completion: Baselines strike back}.
\newblock arXiv preprint arXiv:1705.10744  (2017).

\bibitem{ds:levy}
O.~Levy and I.~Dagan.
\newblock \emph{{Annotating Relation Inference in Context via Question
  Answering.}}
\newblock ACL  (2016).

\bibitem{ds:freebase}
K.~Bollacker, C.~Evans, P.~Paritosh, T.~Sturge, and J.~Taylor.
\newblock \emph{Freebase: a collaboratively created graph database for
  structuring human knowledge}.
\newblock In \emph{Proceedings of the 2008 ACM SIGMOD international conference
  on Management of data}, pp. 1247--1250 (AcM, 2008).

\bibitem{ds:wordnet}
C.~Fellbaum.
\newblock \emph{WordNet} (Wiley Online Library, 1998).

\bibitem{openie}
G.~Angeli, M.~J. Premkumar, and C.~D. Manning.
\newblock \emph{Leveraging linguistic structure for open domain information
  extraction}.
\newblock In \emph{Proceedings of the 53rd Annual Meeting of the Association
  for Computational Linguistics (ACL 2015)} (2015).

\bibitem{riedel2013relation}
S.~Riedel, L.~Yao, A.~McCallum, and B.~M. Marlin.
\newblock \emph{Relation extraction with matrix factorization and universal
  schemas.}
\newblock In \emph{HLT-NAACL}, pp. 74--84 (2013).

\bibitem{ds:nyt}
E.~Sandhaus.
\newblock \emph{The new york times annotated corpus}.
\newblock Linguistic Data Consortium, Philadelphia \textbf{6}(12), e26752
  (2008).

\bibitem{Goldberg:2013wd}
Y.~Goldberg and J.~Orwant.
\newblock \emph{{A Dataset of Syntactic-Ngrams over Time from a Very Large
  Corpus of English Books.}}  (2013).

\bibitem{ngrams}
Y.~Goldberg and J.~Orwant.
\newblock \emph{A dataset of syntactic-ngrams over time from a very large
  corpus of english books.}
\newblock In \emph{* SEM@ NAACL-HLT}, pp. 241--247 (2013).

\bibitem{Riedel:2013ve}
S.~Riedel, L.~Yao, A.~McCallum, and B.~M. Marlin.
\newblock \emph{{Relation Extraction with Matrix Factorization and Universal
  Schemas.}}
\newblock HLT-NAACL  (2013).

\bibitem{Lin}
D.~Lin and P.~Pantel.
\newblock \emph{{Discovery of inference rules for question-answering.}}
\newblock Natural Language Engineering \textbf{7}(04) (2001).

\bibitem{Cover}
J.~Weeds and D.~Weir.
\newblock \emph{A general framework for distributional similarity}.
\newblock In \emph{Proceedings of the 2003 conference on Empirical methods in
  natural language processing}, pp. 81--88 (Association for Computational
  Linguistics, 2003).

\bibitem{BInc}
I.~Szpektor and I.~Dagan.
\newblock \emph{{Learning entailment rules for unary templates}}.
\newblock In \emph{the 22nd International Conference}, pp. 849--856
  (Association for Computational Linguistics, Morristown, NJ, USA, 2008).

\bibitem{berant}
J.~Berant, N.~Alon, I.~Dagan, and J.~Goldberger.
\newblock \emph{{Efficient Global Learning of Entailment Graphs}}.
\newblock Computational Linguistics \textbf{41}(2), 249 (2015).

\bibitem{mikolov2013distributed}
T.~Mikolov, I.~Sutskever, K.~Chen, G.~S. Corrado, and J.~Dean.
\newblock \emph{Distributed representations of words and phrases and their
  compositionality}.
\newblock In \emph{Advances in neural information processing systems}, pp.
  3111--3119 (2013).

\bibitem{goldberg2014word2vec}
Y.~Goldberg and O.~Levy.
\newblock \emph{word2vec explained: deriving mikolov et al.'s negative-sampling
  word-embedding method}.
\newblock arXiv preprint arXiv:1402.3722  (2014).

\bibitem{complex}
T.~Trouillon, J.~Welbl, S.~Riedel, {\'E}.~Gaussier, and G.~Bouchard.
\newblock \emph{Complex embeddings for simple link prediction}.
\newblock In \emph{International Conference on Machine Learning}, pp.
  2071--2080 (2016).

\bibitem{transe}
A.~Bordes, N.~Usunier, A.~Garcia-Duran, J.~Weston, and O.~Yakhnenko.
\newblock \emph{Translating embeddings for modeling multi-relational data}.
\newblock In \emph{Advances in neural information processing systems}, pp.
  2787--2795 (2013).

\bibitem{riedel}
S.~Riedel, L.~Yao, A.~McCallum, and B.~M. Marlin.
\newblock \emph{Relation extraction with matrix factorization and universal
  schemas.}
\newblock In \emph{HLT-NAACL}, pp. 74--84 (2013).

\bibitem{Berant:2011te}
J.~Berant, I.~Dagan, and J.~Goldberger.
\newblock \emph{{Global learning of typed entailment rules}} (Association for
  Computational Linguistics, 2011).

\bibitem{Levy:2014wv}
O.~Levy, I.~Dagan, J.~Goldberger, and I.~Ramat-Gan.
\newblock \emph{{Focused Entailment Graphs for Open IE Propositions.}}
\newblock CoNLL  (2014).

\bibitem{Zeichner:2012wn}
N.~Zeichner, J.~Berant, and I.~Dagan.
\newblock \emph{{Crowdsourcing inference-rule evaluation}} pp. 156--160 (2012).

\bibitem{distmult}
B.~Yang, W.-t. Yih, X.~He, J.~Gao, and L.~Deng.
\newblock \emph{Embedding entities and relations for learning and inference in
  knowledge bases}.
\newblock arXiv preprint arXiv:1412.6575  (2014).

\bibitem{kingma2014adam}
D.~Kingma and J.~Ba.
\newblock \emph{Adam: A method for stochastic optimization}.
\newblock arXiv preprint arXiv:1412.6980  (2014).

\bibitem{dozat2016incorporating}
T.~Dozat.
\newblock \emph{Incorporating nesterov momentum into adam}  (2016).

\end{thebibliography}
